\documentclass[11pt]{article}

\usepackage[table]{xcolor}
\usepackage[final]{acl}

\usepackage{times}
\usepackage{latexsym}

\usepackage[T1]{fontenc}

\usepackage[utf8]{inputenc}

\usepackage{microtype}

\usepackage{inconsolata}
\usepackage{amsmath}
\usepackage{booktabs}
\usepackage{tabularx}
\usepackage{graphicx}
\usepackage{amsthm,amssymb}
\usepackage{mathrsfs}
\usepackage{xspace}
\usepackage{multirow}
\usepackage{tikz}
\usepackage{array}

\newcolumntype{L}[1]{>{\raggedright\arraybackslash}p{#1}}

\usetikzlibrary{positioning, fit, calc, arrows.meta, shapes.geometric}

\newcommand{\up}[1]{\textcolor{red}{\scriptsize$\uparrow$\,#1}}
\newcommand{\down}[1]{\textcolor{blue}{\scriptsize$\downarrow$\,#1}}
\newcommand{\model}{CTD\xspace}

\usepackage{graphicx}

\usepackage{CJKutf8}  

\title{ChunQiuTR: Time-Keyed Temporal Retrieval in Classical Chinese Annals}

\author{
	 \textbf{Yihao Wang\textsuperscript{1}},
	 \textbf{Zijian He\textsuperscript{1}},
	 \textbf{Jie Ren\textsuperscript{2}},
	 \textbf{Keze Wang\textsuperscript{1}},
	\\
	\\
	 \textsuperscript{1}Sun Yat-Sen University,
	 \textsuperscript{2}Shaanxi Normal University,
	\\
	 \small{
		   \textbf{Correspondence:} \href{kezewang@gmail.com}{kezewang@gmail.com}
		 }
	}

\begin{document}
	\begin{CJK}{UTF8}{gbsn} 
		\maketitle
\begin{abstract}
Retrieval shapes how language models access and ground knowledge in retrieval-augmented generation (RAG). In historical research, the target is often not an arbitrary relevant passage, but the exact record for a specific regnal month, where temporal consistency matters as much as topical relevance. This is especially challenging for Classical Chinese annals, where time is expressed through terse, implicit, non-Gregorian reign phrases that must be interpreted from surrounding context, so semantically plausible evidence can still be temporally invalid. We introduce \textbf{ChunQiuTR}, a time-keyed retrieval benchmark built from the \textit{Spring and Autumn Annals} and its exegetical tradition. ChunQiuTR organizes records by month-level reign keys and includes chrono-near confounders that mirror realistic retrieval failures. We further propose \textbf{CTD} (Calendrical Temporal Dual-encoder), a time-aware dual-encoder that combines Fourier-based absolute calendrical context with relative offset biasing. Experiments show consistent gains over strong semantic dual-encoder baselines under time-keyed evaluation, supporting retrieval-time temporal consistency as a key prerequisite for faithful downstream historical RAG. Our code and datasets are available at
\href{https://github.com/xbdxwyh/ChunQiuTR}{\texttt{github.com/xbdxwyh/ChunQiuTR}}.
\end{abstract}

\section{Introduction}

Retrieval is increasingly the interface between language models and the world’s knowledge, most visibly in retrieval-augmented generation (RAG) and search-augmented assistants~\citep{gao2023retrieval,Lewis2020Retrieval}. In such systems, models ground responses in retrieved evidence rather than relying on parametric memory alone. This evidentiary role is central to expert workflows---literature survey, legal and policy analysis, and scientific claim verification---where users care not only about \emph{what} an answer is, but also \emph{where} it comes from~\citep{menick2022teaching}.

\begin{figure}[t]
    \includegraphics[width=\columnwidth]{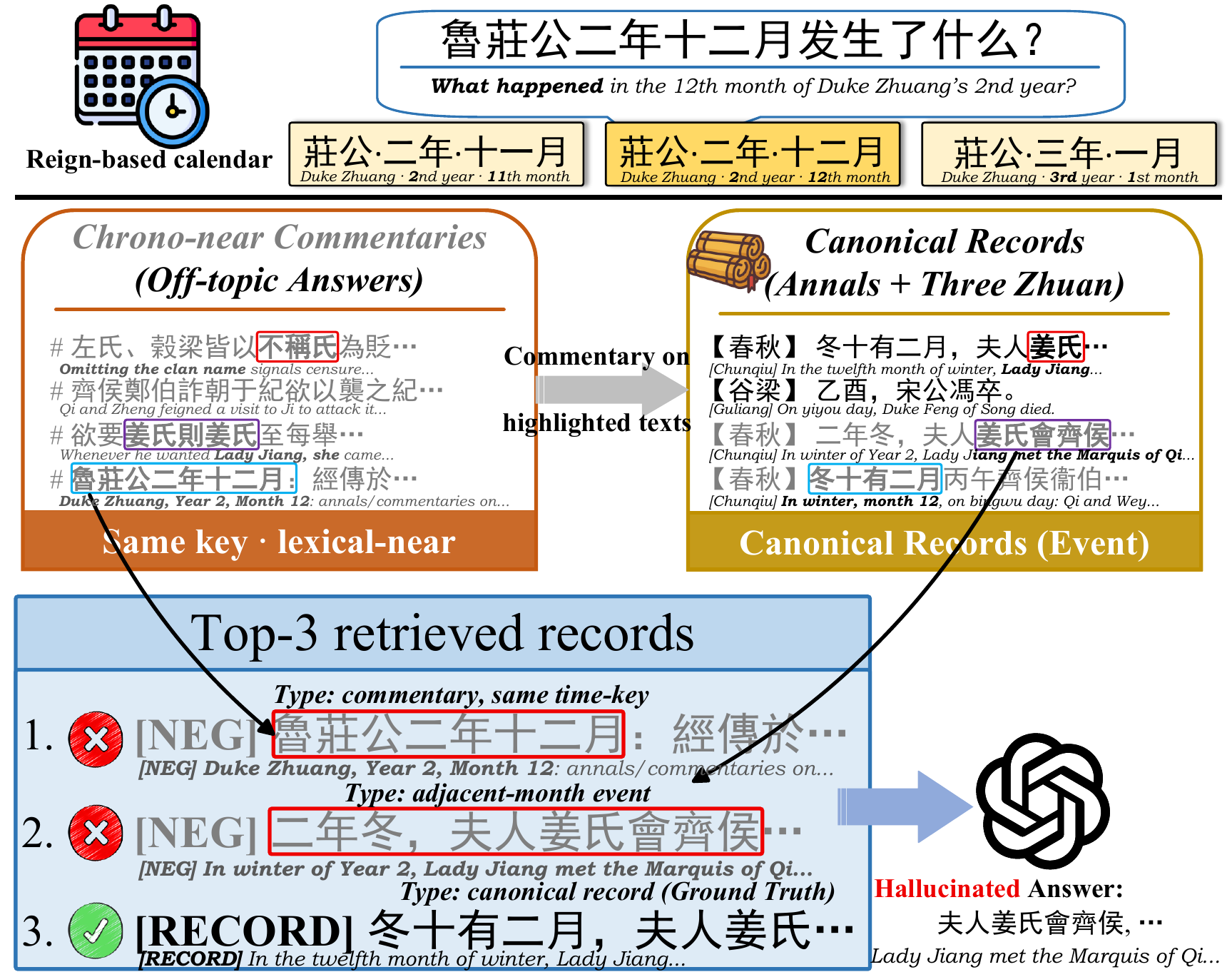}
    \caption{A query about a specific month can retrieve same-month commentary that repeats the date phrase, or adjacent-month near-miss events with confusable wording, so a retrieval-augmented model answers fluently but at the wrong time.}
    \label{fig:intro}
    \vspace{-4.5mm}
\end{figure}

Historical research on pre-modern Chinese sources is a canonical example of evidence-centric retrieval~\citep{cao-etal-2024-tonggu,zhang-etal-2024-philogpt,liu-etal-2025-large}. Digitized annals, commentaries, and later annotations are now searchable, but the target is rarely an arbitrary topical snippet: it is the passage that records what happened in a particular month of a particular duke’s reign. As Fig.~\ref{fig:intro} illustrates, a query such as ``What happened in Duke Zhuang’s 2nd year, 12th month?'' can easily retrieve (i) exegetical commentary that repeats the same date phrase without answering the event, or (ii) near-duplicate events from adjacent months with highly confusable wording. In this setting, semantic relevance is insufficient without verifying temporal alignment to the queried month. Once retrieval binds a downstream generator to temporally incorrect but semantically plausible evidence, the final answer may still sound fluent while being wrong about \emph{when} the event happened.

This motivates a more focused question that is central to faithful historical RAG:

\begin{quote}
\textbf{(Q)} \textit{How can a retriever select \textbf{time-consistent evidence} for queries expressed under \textbf{non-Gregorian}, reign-based chronologies?}
\end{quote}

Studying this problem is already challenging because pre-modern records typically do not provide explicit, globally comparable (Gregorian) timestamps~\citep{chen2021timeqa,Chen2025tempqa}. Instead, they employ a ruler-centric regnal chronology: time is expressed relative to the current ruler and his regnal year and month, so temporal reference effectively resets across reigns and must be interpreted on a corpus-specific timeline rather than a monotonic calendar. Moreover, temporal phrases are often underspecified or written in shorthand---for example, ``in summer, in the fifth month'' may omit the absolute year and only become interpretable given the surrounding reign context. Crucially, time is not a clean metadata field separated from content: in annalistic writing, distinctive one-off events can implicitly function as temporal anchors, tightly coupling \emph{when} with \emph{what}. As a result, retrieval cannot rely on semantic similarity or timestamp ordering alone; it must identify evidence that is both topically relevant and temporally consistent with the intended regnal point or window.

To tackle this challenge, we ground our study in a demanding case: the \textit{Spring and Autumn Annals} and its commentarial--exegetical corpus. We introduce \textbf{ChunQiuTR}, a time-keyed benchmark built on this material, where queries and records are expressed in a ruler-centric, non-Gregorian chronology rather than modern timestamps. Building on this benchmark, we propose the \textbf{Calendrical Temporal Dual-encoder (CTD)}, a time-aware dual-encoder retriever that augments semantic matching with learned calendrical structure. CTD places each query and record at a soft location on a unified ordered calendar axis and favors pairs that agree not only in meaning but also in calendrical position. Concretely, it injects an absolute calendrical context into embeddings and adds a relative temporal bias to similarity based on signed calendar offsets, improving robustness to adjacent-month and lexical-near confounders.

Our contributions are threefold: (i) we introduce \textbf{ChunQiuTR}, a non-Gregorian, reign-keyed temporal retrieval benchmark with point/gap/window queries and leak-free splits; (ii) we propose \textbf{CTD}, a calendrically time-aware dual-encoder that combines absolute context injection with relative offset biasing; and (iii) we show consistent improvements over strong semantic dual-encoder baselines, especially under chrono-near and adjacent-month confounders, supporting the view that retrieval-time temporal consistency is a key prerequisite for faithful downstream historical RAG.

		\section{Related Work}
		
		\subsection{Neural Information Retrieval}
		Lexical retrievers such as BM25~\citep{Robertson2009BM25} remain strong and interpretable, but are limited by surface-form overlap.
		Neural IR methods are often grouped into neural sparse expansion (e.g., SPLADE~\citep{Formal2021SPLADE}), dense dual-encoders trained with contrastive learning (e.g., DPR~\citep{karpukhin-etal-2020-dense}, ANCE~\citep{xiong2021approximate}, Contriever-style~\citep{lei-etal-2023-unsupervised}), and late-interaction token matching (e.g., ColBERT~\citep{Khattab2020ColBERT}).
		General-purpose embedding models (e.g., GTR~\citep{ni-etal-2022-large-GTR}, E5~\citep{wang2024multilinguale5textembeddings}, Qwen3-Embedding~\citep{zhang2025qwen3embeddingadvancingtext}) further enable plug-and-play retrieval.
		However, relevance is typically modeled as semantic similarity, which can still confuse chrono-near near-duplicates or fail to enforce fine-grained temporal constraints without explicit temporal structure.
		
		\subsection{Temporal Information Retrieval}
		Temporal information retrieval (TIR) incorporates time into ranking, ranging from timestamp-aware priors (e.g., time-based language models~\citet{li2003timebasedlm}) to modern formulations emphasizing temporal focus/intent~\citep{piryani2025itshightimesurvey}.
		Recent work explores neural retrieval for time-sensitive settings by injecting temporal signals into retrieval or generation pipelines~\citep{Rajapakse2023densePassage,zhang2025mrag}, as well as mechanisms that encode time specifiers into model behaviors~\citep{han-etal-2025-temporal}.
		Related lines include time-aware language models with document-dating objectives and temporal label-smoothing schemes that smooth supervision over neighboring time steps, both of which we echo in our auxiliary temporal heads~\citep{wang2023BiTimeBERT,pmlr-v202-yeche23a,dhingra-etal-2022-time}. Most TIR studies target modern timestamped collections under open retrieval, whereas our setting is a micro-granular, \emph{time-keyed} chronicle with dense chrono-near near-duplicates, leading to different supervision and evaluation objectives. 
        
        Temporal-expression extraction and normalization are also related to our setting, including work on historical texts and cross-lingual temporal expression extraction/normalization~\citep{Korchagina2016,Cao2022XLTime,Su2025TemporalIEReview,SanchezDeCastro2025TemporalNormalization,Graciotti2025KEMHISTO}. 
        However, in many Chunqiu passages, the ruling duke and/or regnal year is omitted, so the target month key must be recovered from annalistic structure and discourse context rather than extracted as a standalone temporal mention.

		\section{Dataset Construction}

\begin{figure*}[t]
    \includegraphics[width=\linewidth]{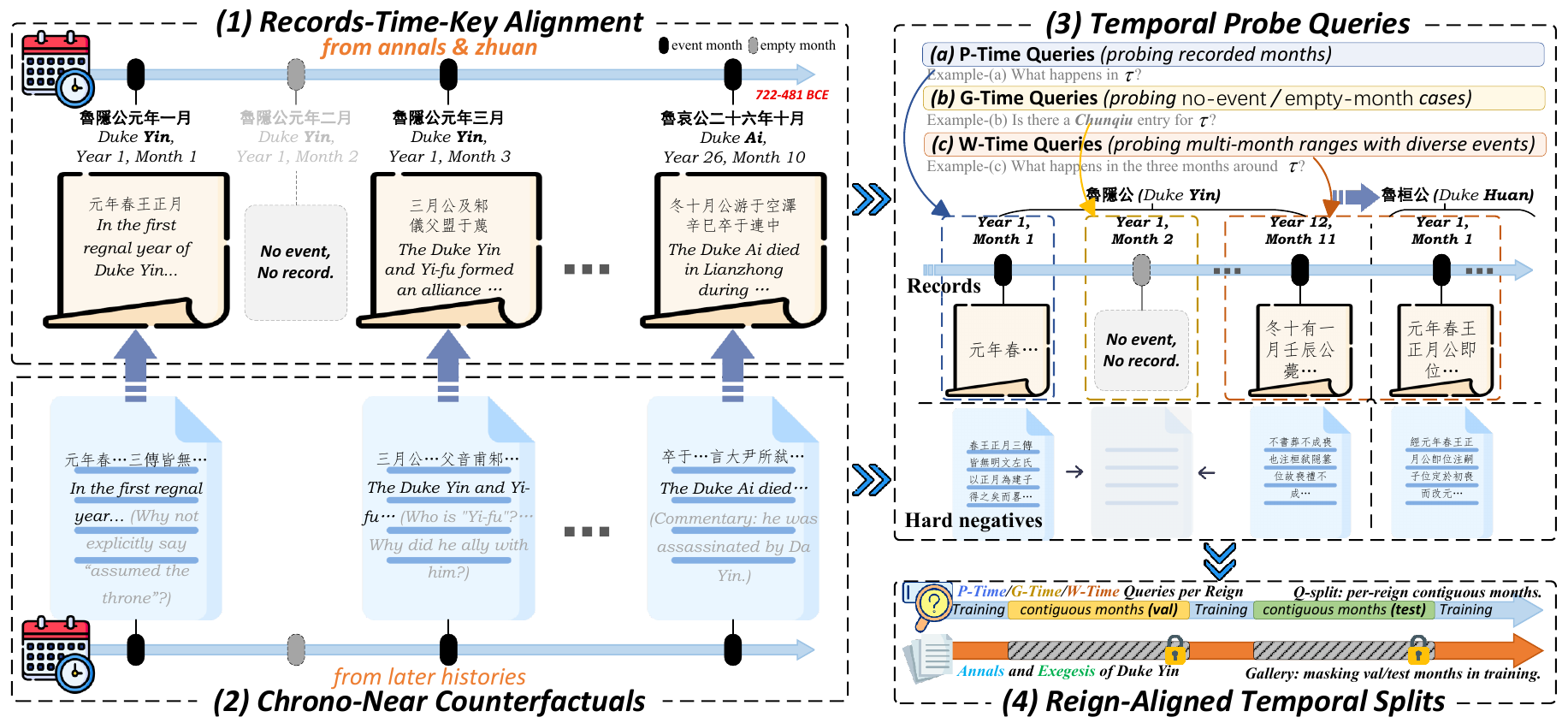}
    \caption{Overview of the ChunQiuTR construction. (\textbf{Left}) Time-key alignment yields event-level records, augmented with chrono-near counterfactual hard negatives from later histories. (\textbf{Right}) We define P/G/W temporal queries and leak-free, reign-aligned splits.}
    \label{fig:data_pipeline}
    \vspace{-4.5mm}
\end{figure*}

To study temporal retrieval under non-Gregorian dating systems, we construct \textbf{ChunQiuTR}, a benchmark curated from authentic historical texts centered on the \textit{Spring and Autumn Annals} and its classical commentarial tradition. 
The retrieval gallery is derived from source texts rather than AI-generated content, and the queries are instantiated from a small set of manually written templates. 
We use LLMs only as auxiliary tools to propose candidate splits or candidate alignments during curation; they are never used to generate, rewrite, translate, or paraphrase historical content, and only human-approved results enter the final benchmark. 
ChunQiuTR evaluates whether models can identify both \emph{events} and their \emph{temporal positions} when time is expressed using reign-year references rather than explicit Gregorian dates. Fig.~\ref{fig:data_pipeline} outlines the data construction process. We next explain why the \textit{Chunqiu} is a natural testbed and how the benchmark is instantiated.

\subsection{Chunqiu Corpus and Temporal Scheme}
\label{sec:time_key}

\paragraph{Why the \textit{Chunqiu}?}
The \textit{Chunqiu} (《春秋》, \emph{Spring and Autumn Annals}) is a terse chronicle of the state of Lu (722--481~BCE) with a long exegetical and historiographical tradition built around it. Its entries are extremely compact and date events only using a reign-based temporal language. Brief formulae such as ``元年春'' (first year, spring) or ``夏五月'' (summer, fifth month) omit explicit absolute years and often leave the ruler implicit. Because many landmark events occur only once, mentioning the event together with such a relative month phrase often suffices to pin down a unique point on the timeline. These compressed records are then expanded and re-interpreted by the three classical \textit{zhuan} (\textit{Zuo}, \textit{Gongyang}, \textit{Guliang}) and later commentarial and historiographical works.

Taken together, this layered structure makes the \textit{Chunqiu} corpus an ideal testbed for temporal retrieval: all layers align to the same ruler-centric timeline without explicit Gregorian dates, and different layers often describe the same event or nearby months with overlapping phrasing, naturally producing realistic ``near-miss'' hard negatives. Similar combinations of a shared chronological backbone and dense, overlapping commentary recur in many pre-modern annalistic corpora, so we use the \textit{Chunqiu} as a compact starting point for methods that can later be generalized to broader non-Gregorian historical collections. Additional details on sources and preprocessing are provided in Appendix~\ref{sec:appendix_detail_data}.

\paragraph{Reign-based time keys.}
Unlike modern texts that use absolute year numbering (e.g., ``709~BCE''), our sources use a ruler-centric, reign-based dating scheme in which regnal years restart for each new Lu duke. A regnal year may begin with a full formula such as ``元年春王正月'' (first year, spring, royal first month), but later entries are often shortened to month phrases like ``夏五月'' (summer, fifth month), with the duke and year supplied by context---effectively ``in the fifth month of summer of this duke's current regnal year.'' Because many key events in the \textit{Chunqiu} are unique, the event description plus such a relative month phrase often pins down a single point on the reign-based timeline. This compact temporal language contrasts with standard time-IR corpora, where documents carry explicit Gregorian dates that can be treated as fixed metadata rather than temporal expressions to be interpreted.

Importantly, assigning a month-level time key in this corpus is not reducible to standard temporal-expression extraction. The ruling duke and/or regnal year is often omitted and must be recovered from annalistic structure, discourse continuity, and surrounding context. We therefore manually verify the final time-key assignment for all records rather than relying on fully automatic extraction.

For our benchmark, we normalize this temporal language into month-level time keys~\(\tau = (\text{gong}, \text{year}, \text{month})\) to index all records, where \(\text{gong}\) denotes the ruling duke title (Chinese character ``公'', glossed as ``Duke'' for readability). Month-level is the finest temporal unit consistently recorded in the \textit{Chunqiu}; finer-grained dates (e.g., day-level) are largely absent or sporadic and thus cannot be normalized systematically. We assign a time key to every month, including months with no annals entry; we later instantiate them as standardized \texttt{no\_event} placeholders. We then align each annals entry and its associated historiographical passages to a unique \(\tau\), forming a time-keyed gallery of short records (Section~\ref{sec:record_alignment}, Appendix~\ref{sec:reign-time-key-details}). Although in the \textit{Chunqiu} this appears as a concrete \(\text{gong}\)--\(\text{year}\)--\(\text{month}\) scheme, the pipeline itself only assumes an ordered set of time keys with aligned texts and thus applies to other non-Gregorian chronologies.

\subsection{Record Alignment}
\label{sec:record_alignment}

\paragraph{Record--time-key alignment.}
Building on the reign-based month keys $\tau=(\text{gong},\text{year},\text{month})$ defined above, we define the record units used for retrieval as shown in Fig.~\ref{fig:data_pipeline}(1). We treat each \emph{record} as the atomic retrieval unit: a short event-level passage aligned to a single month key~$\tau$. For each time key~\(\tau\), we gather all snippets from the annals and the three classical \textit{zhuan} and refine them into event-level record sets \(\mathcal{D}_{\tau}\). This is non-trivial: the annals often compress several events into a single sentence, while the commentaries may spread one event across multiple fragments, so naive sentence or paragraph boundaries do not match historical events. We therefore use a lightweight LLM prompt only to propose candidate splits and groupings under each \(\tau\), and then manually review and correct these proposals so that each final record corresponds to one coherent event with its aligned commentarial material. This yields a high-precision, time-keyed collection of event-level records for the entire \textit{Chunqiu} period (Appendix~\ref{sec:appendix_record_alignment}).

\paragraph{Chrono-near counterfactual negatives.}
Later historiographical layers, such as Gu Donggao's \emph{Chronological Tables}, re-group and paraphrase the same Spring and Autumn events instead of introducing new ones, creating naturally occurring, easy-to-confuse off-topic variants. As shown in Fig.~\ref{fig:data_pipeline}(2), we align these sources to our reign-based time keys using LLM-assisted candidate matching together with fuzzy string matching (Appendix~\ref{sec:appendix_counterfactual}). Here again, the LLM is used only to propose candidate source passages; final alignments are retained only after human verification. For each time key~\(\tau\), we define \(\mathcal{D}^{\text{cf}}_{\tau}\) as records from these layers that share the same time key and describe the same situation in later paraphrase, but are \emph{not used as ground-truth retrieval targets} for queries targeting~$\tau$. These historically grounded near-miss variants serve as hard negatives for time-aware retrieval.

\paragraph{Audit and reliability.}
We summarize three reliability decisions here and defer full details to the appendix. First, time-key normalization is manually verified because many passages omit an explicit duke or regnal year and must be resolved from annalistic structure and discourse continuity rather than extracted as standalone temporal mentions. Second, among 1{,}533 non-empty months, 558 contain multiple events; after LLM candidate grouping, only 63 required additional human correction, while the remaining 495 were accepted without change. Third, for later-commentary alignment, LLMs only propose candidate matched passages, and the final human acceptance rate ranges from 93.33\% to 100\% across sources. These audits indicate that LLM assistance improves curation efficiency, while final benchmark quality is controlled by explicit human verification (Appendix~\ref{sec:appendix_alignment_audit}).

\subsection{Temporal Queries and Evaluation Splits}
\label{sec:query_design}

\paragraph{Temporal query design.}
Building on the time-keyed records \(\{\mathcal{D}_{\tau}\}\), we design three families of temporal queries that mirror common ways historians ask about time: point queries (\textbf{P-Time}), gap queries (\textbf{G-Time}), and local-window queries (\textbf{W-Time}) as shown in Fig.~\ref{fig:data_pipeline}(3). Each query is mapped to a target interval \(Q_i\) on the reign-based timeline; span (single month vs.\ short multi-month window) and eventfulness (whether \(Q_i\) contains empty months) vary independently, so any family can in principle target either eventful or empty periods.

To make gaps and empty intervals queryable, months with no annals entry are instantiated as standardized \texttt{no\_event} records keyed by \(\tau\) and included in the retrieval gallery. G-Time queries are gap-oriented: they ask which month(s) within a reign or specified range lack recorded events and are answered by these \texttt{no\_event} records.

P-Time queries explicitly target a single time key (e.g., ``What happens in \emph{鲁隐公元年三月}?''), and are answered by the corresponding event-bearing record, or by an explicit \texttt{no\_event} record when that month is empty. W-Time queries target a local temporal window, such as ``around the time when \(X\) occurs'' or ``in the months before/after \(Y\),''
and are mapped to short contiguous ranges of time keys (see Fig.~\ref{fig:app_temporal_reasoning_patterns} for representative patterns). All query templates and concrete examples are provided in Appendix~\ref{sec:appendix_queries}.

\paragraph{Reign-aligned splits.}
We partition the month-level timeline into train/validation/test splits in a reign-aware, leak-free way. For each duke's reign, we allocate disjoint contiguous blocks of months to train, validation, and test, roughly in an 80/10/10 ratio, and assign all records and queries in those months to the corresponding split. No time key, record, or query ever appears in more than one split. At evaluation time, validation and test queries are drawn from their held-out reign segments, while retrieval is always performed over the full time-keyed record gallery, so models must generalize temporal reasoning from seen parts of a reign to months they have \emph{never} observed during training. Overall, the benchmark contains 20{,}172 records and 16{,}226 queries (13{,}053 train / 1{,}520 validation / 1{,}653 test); detailed statistics and split visualizations are provided in Appendix~\ref{sec:appendix_statistic_month}.

        \section{Methods}
\begin{figure*}[t]
			\includegraphics[width=\linewidth]{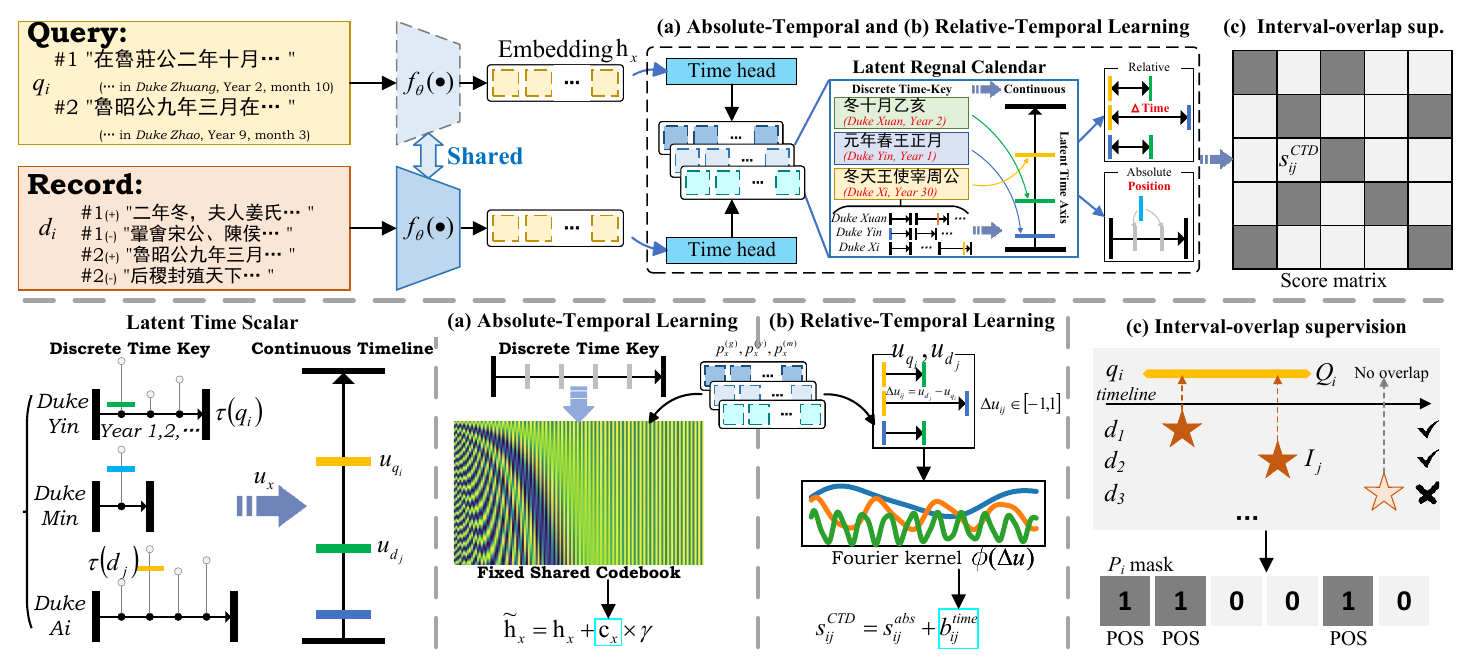}
			\caption{
            Overview of our Calendrical Temporal Dual-encoder (CTD).
            A shared Transformer dual-encoder encodes queries and records into embeddings.
            Temporal heads place each text on a unified regnal calendar axis (latent time scalar), supporting (a) absolute context injection and (b) relative biasing to form \(s^{\text{CTD}}_{ij}\).
            (c) Interval-overlap supervision marks in-batch multi-positives by query–record overlap and trains a multi-positive contrastive loss.
            }

			\label{fig:method_overview}
            \vspace{-4.5mm}
		\end{figure*}

        We first formalize ChunQiuTR as a time-keyed retrieval task in Sec.~\ref{sec:task}.
        Sec.~\ref{sec:ctd_temporal_encoder} then presents our Calendrical Temporal Dual-encoder (CTD): starting from a semantic dual-encoder score, CTD learns a latent regnal calendar scalar and incorporates an \textbf{absolute} calendrical context and a \textbf{relative} temporal bias to form the final score, as illustrated in Fig.~\ref{fig:method_overview}.
        Finally, Sec.~\ref{sec:loss_function} describes our interval-overlap multi-positive supervision and joint training objective.

        \subsection{Task Formulation}
        \label{sec:task}
        
        We cast our ChunQiuTR benchmark as a temporal retrieval task over a discrete, reign-based month timeline.
        Building on the time-keyed records in Sec.~\ref{sec:record_alignment}, we formalize all aligned historical material as a fixed retrieval gallery
        \[
        \mathcal{D} = \{ d_j \}_{j=1}^{N},
        \]
        where each short Classical Chinese record \(d_j\) is associated with a reign-based month key \(\tau(d_j)\); we write
        \(\mathcal{D}_{\tau} = \{ d_j \in \mathcal{D} : \tau(d_j) = \tau \}\) for the subset under time key \(\tau\).
        
        For each query \(q_i\) constructed in Section~\ref{sec:query_design}, the benchmark specifies a target interval \(Q_i\) on the same month axis and a small multi-positive ground-truth set
        \[
        \mathcal{G}_i \subseteq \mathcal{D}.
        \]
        Ground-truth records \(d_j \in \mathcal{G}_i\) are exactly those that describe events or explicit non-events recorded during the queried interval \(Q_i\), i.e., their time keys \(\tau(d_j)\) fall within \(Q_i\). The learning objective is to train a scoring function \(S_{\theta}(q_i, d_j)\) that, for each query, ranks its ground-truth set \(\mathcal{G}_i\) ahead of the remaining elements of \(\mathcal{D}\).

        \subsection{CTD: Calendrical Temporal Dual-encoder}
        \label{sec:ctd_temporal_encoder}
        We instantiate \(S_{\theta}\) with a standard dual-encoder retriever. A shared Transformer encoder \(f_\theta(\cdot)\) maps both temporal queries and candidate records into a common embedding space, producing pooled embeddings \(\mathbf{h}_{q_i}, \mathbf{h}_{d_j} \in \mathbb{R}^H\).
        As a \emph{purely semantic} baseline, we compute temperature-scaled dot-product similarities
        \(
        s^{\text{sem}}_{ij} = s^{\text{sem}}(q_i,d_j) = \mathbf{h}_{q_i}^\top \mathbf{h}_{d_j} / \alpha,
        \)
        for a mini-batch of $B$ queries $\{q_i\}_{i=1}^B$ and $B$ records $\{d_j\}_{j=1}^B$.
        Building on this semantic score, CTD augments the retriever with (i) an \emph{absolute} calendrical context injected into the embeddings and (ii) a \emph{relative} temporal bias added to the similarity, so that matches must agree in both meaning and calendrical position.

        \subsubsection{Latent calendar scalar}
        
        \label{sec:time_head}
        Reign-based month keys are discrete identifiers and do not directly provide a metric notion of \emph{position} or \emph{distance} across the stitched regnal calendar.
        To support both absolute positioning (for context injection) and relative offsets (for biasing), we therefore learn a continuous calendar axis where temporal relations become measurable.
        
        For any text \(x\) (either a query \(q_i\) or a record \(d_j\)), let \(\mathbf{h}_x \in \mathbb{R}^H\) denote its pooled embedding.
        On top of \(\mathbf{h}_x\), we attach three lightweight prediction heads for \emph{gong}, \emph{year}, and \emph{month}.
        Each head produces logits over its discrete index set, which we normalize into distributions
        \(\mathbf{p}^{(g)}_x, \mathbf{p}^{(y)}_x, \mathbf{p}^{(m)}_x\).
        Taking expectations yields soft calendrical coordinates \(g_x, y_x, m_x\), which locate \(x\) on the ruler–year–month grid.
        
        We then linearize this grid in calendar order and normalize it to \([0,1]\), defining a shared latent time scalar
        \[
        u_x
        = \frac{g_x \cdot (Y \cdot M) + y_x \cdot M + m_x}{G \cdot Y \cdot M - 1}
        \in [0,1].
        \]
        Here \(G\), \(Y\), and \(M\) denote the (padded) maximum numbers of gongs, years-per-gong, and months-per-year used to index the unified calendar.
        Texts from earlier dukes, years, or months receive smaller \(u_x\) than those later in the chronicle, enabling both relative distances \(\Delta u\) and absolute positions to be modeled on the same axis.

        \subsubsection{Absolute-temporal learning}
        \label{sec:absolute_time}
        We first exploit this signal in an \emph{absolute} manner (Fig.~\ref{fig:method_overview}~(a)): instead of feeding discrete $(\text{gong},\text{year},\text{month})$ indices as hard metadata, we convert the heads' \emph{soft} predictions into a continuous context vector and inject it into the embedding.
        
        Reusing $\mathbf{p}^{(g)}_x, \mathbf{p}^{(y)}_x, \mathbf{p}^{(m)}_x$, we map each calendrical index to a fixed Fourier-style code and build sinusoidal codebooks
        \[
        E^{(g)} \in \mathbb{R}^{G \times D_t},\;
        E^{(y)} \in \mathbb{R}^{Y \times D_t},\;
        E^{(m)} \in \mathbb{R}^{M \times D_t}.
        \]
        This fixed sinusoidal codebook provides a smooth, non-parametric absolute-position signal, avoiding a large learned embedding table for sparse calendrical indices.
        Taking expectations under $\mathbf{p}_x^{(\cdot)}$ yields a mixture representation that naturally reflects the model's uncertainty instead of committing to a single hard index.

        We obtain soft absolute-time contexts by taking expectations:
        \[
        \mathbf{c}^{(g)}_x = \mathbf{p}^{(g)}_x E^{(g)},
        \mathbf{c}^{(y)}_x = \mathbf{p}^{(y)}_x E^{(y)},
        \mathbf{c}^{(m)}_x = \mathbf{p}^{(m)}_x E^{(m)}.
        \]
        Concatenating and projecting yields
        \[
        \mathbf{c}_x = W_{\text{ctx}}\bigl[\mathbf{c}^{(g)}_x;\mathbf{c}^{(y)}_x;\mathbf{c}^{(m)}_x\bigr]\in\mathbb{R}^H,
        \]
        which we inject via a scalar-gated residual
        \[
        \tilde{\mathbf{h}}_x = \mathbf{h}_x + \gamma \mathbf{c}_x,
        \]
        where $\gamma$ is learned.
        
        We compute similarities with the context-enriched representations,
        \[
        s^{\text{abs}}_{ij} = \tilde{\mathbf{h}}_{q_i}^\top \tilde{\mathbf{h}}_{d_j} / \alpha,
        \]
        which reduces to the semantic baseline $s^{\text{sem}}_{ij}$ when $\gamma=0$.

        \subsubsection{Relative-temporal learning.}
        \label{sec:relative_time}
        Building on the absolute similarity $s^{\text{abs}}_{ij}$, we further use the learned calendar axis to bias matching by relative offsets (Fig.~\ref{fig:method_overview}~(b)).
        Given the latent coordinates $u_{q_i}$ and $u_{d_j}$ for a query--record pair $(q_i,d_j)$, we form the temporal offset
        \[
        \Delta u_{ij} = u_{d_j} - u_{q_i} \in [-1,1],
        \]
        so that distances along the learned timeline can modulate how easily two texts should match.
        We embed this scalar with Fourier-style features
        \[
        \phi(\Delta u_{ij}) \in \mathbb{R}^{D_\phi},
        \]
        and apply a small MLP to produce an additive temporal bias
        \[
        b^{\text{time}}_{ij} = \epsilon\,\mathrm{MLP}\bigl(\phi(\Delta u_{ij})\bigr).
        \]
        The final retrieval score is
        \[
        s^{\text{CTD}}_{ij} = s^{\text{abs}}_{ij} + b^{\text{time}}_{ij},
        \]
        where the learnable scale $\epsilon$ (initialized near zero) keeps the bias lightweight: when $\epsilon=0$, CTD reduces to the absolute-only scorer $s^{\text{abs}}_{ij}$, and more generally the model can downweight this term if the learned calendar signal is unreliable.

        \subsection{Learning Objectives}
        \label{sec:loss_function}
        We train the purely semantic dual-encoder baseline with a symmetric single-positive InfoNCE~\citep{pmlr-v119-chen20j} objective over $s^{\text{sem}}_{ij}$. For CTD, we instead optimize a temporally aware multi-positive retrieval loss using the final scores $s^{\text{CTD}}_{ij}$.
        
        \paragraph{Interval-overlap multi-positive retrieval.}
        As shown in Fig.~\ref{fig:method_overview} (c), we treat temporal overlap as weak supervision: each query $q_i$ targets an interval $Q_i=[\tau_i^{\min},\tau_i^{\max}]$, and each record $d_j$ carries a single month key $\tau(d_j)$ (i.e., $I_j=[\tau(d_j),\tau(d_j)]$).
        We mark in-batch positives by overlap,
        \[
        P_i=\{\,j \mid Q_i \cap I_j \neq \emptyset \,\},
        \]
        and optimize a multi-positive InfoNCE loss using the final scores $s^{\text{CTD}}_{ij}$:
        \[
        \mathcal{L}_{q}^{\text{multi}}
        = -\frac{1}{B} \sum_{i=1}^B
        \log
        \frac{\sum_{j \in P_i} \exp\!\left(s^{\text{CTD}}_{ij}\right)}
             {\sum_{k=1}^B \exp\!\left(s^{\text{CTD}}_{ik}\right)}.
        \]
        The remaining in-batch records serve as negatives. We define $\mathcal{L}_{d}^{\text{multi}}$ symmetrically by transposing $\bigl(s^{\text{CTD}}_{ij}\bigr)$ and use
        \[
        \mathcal{L}_{\text{multi}}=\tfrac{1}{2}\bigl(\mathcal{L}_{q}^{\text{multi}}+\mathcal{L}_{d}^{\text{multi}}\bigr).
        \]

        \paragraph{Auxiliary calendrical classification.}
        To stabilize the absolute calendrical signal, we supervise the gong/year/month heads on passages with cross-entropy:
        \[
        \mathcal{L}_{\text{time}}
        = \mathbb{E}_{d \sim \text{batch}}
        \bigl[\textstyle \sum_{r \in \{g,y,m\}} \mathrm{CE}\!\left(\mathbf{p}^{(r)}_{d},\, y^{(r)}_{d}\right)\bigr]
        \]
        where $y^{(r)}_{d}$ are the ground-truth calendrical labels from the aligned time keys (queries are unlabeled).

        \paragraph{Overall objective.}
        We jointly optimize the retrieval and auxiliary temporal losses with a small weight $\lambda_{\text{time}}$:
        \[
        \mathcal{L}_{\text{total}}
        = \mathcal{L}_{\text{multi}}
        + \lambda_{\text{time}} \,\mathcal{L}_{\text{time}}.
        \]

		\section{Experiments}
        \subsection{Experiment Setting}
        We fine-tune two dual-encoder backbones, \textsc{BERT-base-chinese} and \textsc{Qwen3-Embed-0.6B}, on ChunQiuTR.
        Model details are deferred to Appendix~\ref{app:model_details}, training cost and compute settings to Appendix~\ref{app:compute}, and the full list of compared methods to Appendix~\ref{sec:app_details_comp_baseline}.

		\subsection{Main Results}
		
		\begin{table*}[t]

			\centering
			\small
            \resizebox{0.95\linewidth}{!}{%
			\setlength{\tabcolsep}{4pt}        
			\begin{tabular}{l c l c c c c c}
				\toprule
				\textbf{Method} & \textbf{Pub} & \textbf{Type} &
				\textbf{R@1} & \textbf{R@5} & \textbf{R@10} & \textbf{MRR@10} & \textbf{nDCG@10} \\
				\midrule
				
				\rowcolor{gray!15}\multicolumn{8}{c}{\textbf{\textit{Sparse retrieval}}} \\
				BM25                          & --        & Sparse & 0.3962 & 0.5209 & 0.5620 & 0.4487 & 0.3404 \\
				BM25+TimeKDE				  & --        & Sparse & 0.4943 & 0.6086 & 0.6709  & 0.5456 & 0.4222 \\
				SPLADE-IDF$_{(\text{ZS})}$    & arXiv'24  & Sparse & 0.1361 & 0.2765 & 0.3569 & 0.1971 & 0.1596 \\
				SPLADE-$\ell_0$$_{(\text{ZS})}$ & SIGIR'25 & Sparse & 0.0006 & 0.0309 & 0.0587 & 0.0132 & 0.0143 \\
				
				\midrule
				\rowcolor{gray!15}\multicolumn{8}{c}{\textbf{\textit{Fusion / late interaction}}} \\
				
				ColBERT-JINA$_{(\text{ZS})}$  & MRL'24       & Fusion & 0.2498 & 0.4102 & 0.4743 & 0.3167 & 0.2569 \\
				ColBERT-LFM2$_{(\text{ZS})}$  & arXiv'25     & Fusion & 0.3345 & 0.4567 & 0.4894 & 0.3865 & 0.2691 \\
				
				\midrule
				\rowcolor{gray!15}\multicolumn{8}{c}{\textbf{\textit{Dense retrieval (encoder-based)}}} \\
				mE5-Large$_{(\text{ZS})}$          & arXiv'24        & Dense & 0.2916 & 0.3969 & 0.4574 & 0.3389 & 0.2441 \\
				mE5-Large-ins$_{(\text{ZS})}$      & arXiv'24        & Dense & 0.2359 & 0.3545 & 0.4162 & 0.2862 & 0.2358 \\
				GTE-Large$_{(\text{ZS})}$          & arXiv'23        & Dense & 0.2293 & 0.3527 & 0.3890 & 0.2826 & 0.2188 \\
				BGE-Large-v1.5$_{(\text{ZS})}$     & arXiv'23        & Dense & 0.2208 & 0.3430 & 0.4144 & 0.2775 & 0.2280 \\
				BGE-m3$_{(\text{ZS})}$             & Findings ACL'24 & Dense & 0.2698 & 0.3775 & 0.4253 & 0.3135 & 0.2299 \\
				BERT-base$_{(\text{FT})}$          & NAACL'19        & Dense & 0.5088 & 0.6279 & 0.6727 & 0.5597 & 0.4283 \\
				BERT-base + TempDate$_{(\text{FT})}$   & SIGIR'23        & Dense & 0.5027 & 0.6165 & 0.6691 & 0.5508 & 0.4243 \\
				BERT-base + TempDate-Smooth$_{(\text{FT})}$      & PMLR'23         & Dense & 0.5051 & 0.6152 & 0.6673 & 0.5519 & 0.4244 \\
				\textbf{\model$_{\text{BERT-base}}$ (Ours)} &
				This work        & Dense & \textbf{0.5826} & \textbf{0.6721} & \textbf{0.7090} & \textbf{0.6193} & \textbf{0.4575} \\
				
				
				\midrule
				\rowcolor{gray!15}\multicolumn{8}{c}{\textbf{\textit{Dense retrieval (LM-based embeddings)}}} \\
				GTE-Qwen2-1.5B$_{(\text{ZS})}$     & arXiv'23        & Dense & 0.2783 & 0.4453 & 0.5009 & 0.3501 & 0.2613 \\
				E5-mistral-7B$_{(\text{ZS})}$      & ACL'24          & Dense & 0.2196 & 0.3212 & 0.3684 & 0.2619 & 0.2359 \\
				PQR (Qwen2.5-7B)$_{(\text{re})}$   & ACL'25          & Dense & 0.1585 & 0.3134 & 0.3805 & 0.2226 & 0.1712 \\
				PQR (Qwen3-8B)$_{(\text{re})}$     & ACL'25          & Dense & 0.0901 & 0.2184 & 0.3152 & 0.1481 & 0.1184 \\
				Qwen3-Embed-0.6B$_{(\text{ZS})}$   & arXiv'25        & Dense & 0.3376 & 0.4852 & 0.5354 & 0.3973 & 0.3107 \\
				Qwen3-Embed-4B$_{(\text{ZS})}$     & arXiv'25        & Dense & 0.4410 & 0.5783 & 0.6013 & 0.4985 & 0.3793 \\
				Qwen3-Embed-0.6B$_{(\text{FT})}$   & arXiv'25        & Dense & 0.5771 & 0.6376 & 0.6818 & 0.6045 & 0.4460 \\
				Qwen3-Embed-0.6B + TempDate$_{(\text{FT})}$    & SIGIR'23        & Dense & 0.5523  & 0.6425  & 0.6630   & 0.5924   & 0.4391\\
				Qwen3-Embed-0.6B + TempDate-Smooth$_{(\text{FT})}$      & PMLR'23         & Dense & 0.5638  & 0.6346  & 0.6727   & 0.5942   & 0.4396 \\
				\textbf{\model$_{\text{Qwen3-Embed-0.6B}}$ (Ours)} &
				This work        & Dense & \textbf{0.5923} & \textbf{0.6485} & \textbf{0.6927} & \textbf{0.6194} & \textbf{0.4575} \\
				\bottomrule
			\end{tabular}
            }
			
        \caption{
        Test-set retrieval performance on our ChunQiuTR benchmark under the official evaluation protocol. 
        \label{tab:main_results}
        }
        \vspace{-3mm}
		\end{table*}
		
        From Table~\ref{tab:main_results}, ChunQiuTR is clearly non-trivial and strongly time-sensitive: most zero-shot sparse, fusion, and dense retrievers lag behind tuned BM25, while a simple temporal prior (BM25+TimeKDE) yields a large gain over BM25 and nearly matches supervised dense models. On encoder-based dense retrievers, in-domain fine-tuning already outperforms BM25+TimeKDE, generic dating auxiliaries (TempDate / TempDate-Smooth) give little benefit, and adding our CTD objectives on the same backbone yields a clear boost in early precision (around +7–8 points on R@1). LM-based dense retrievers show a similar pattern: zero-shot LM encoders and PQR pipelines underperform BM25+TimeKDE, lightly fine-tuned Qwen3-Embed-0.6B is strong, and the CTD-enhanced variant further improves early precision and achieves the best overall scores (R@1, MRR@10, nDCG@10), indicating that explicit time-key supervision adds fine-grained temporal structure beyond simple priors or auxiliary dating heads.

\paragraph{Cross-corpus pilot on \textit{Zizhi Tongjian}.}

As an out-of-domain probe, we further evaluate ChunQiuTR-trained retrievers on two processed subsets from \textit{Zizhi Tongjian}, an annalistic general history that also records events under non-Gregorian, reign-based temporal expressions (Appendix~\ref{app:zztj-pilot}). 
For each subset, we build a month-level gallery from event-bearing lines and automatically instantiate one point-style query for each unique month key using traditional reign-year expressions, without any additional training on the target corpus. 
Unlike the full ChunQiuTR benchmark, this pilot does not reconstruct explicit \texttt{no\_event} months, commentary-derived hard negatives, or the full point/gap/window query families, and is therefore intended as a lightweight cross-corpus transfer probe rather than a second benchmark.

\begin{table}[t]
\centering
\small
\setlength{\tabcolsep}{4pt}
\resizebox{\columnwidth}{!}{
\begin{tabular}{lcc|cc|cc}
\toprule
\multirow{2}{*}{Subset} & \multirow{2}{*}{Records} & \multirow{2}{*}{Queries}
& \multicolumn{2}{c|}{FT baseline} & \multicolumn{2}{c}{CTD (ours)} \\
& & & MRR & R@1 & MRR & R@1 \\
\midrule
Qi Ji (part)  & 268 & 92  & 0.2081 & 0.1848 & \textbf{0.2304} & \textbf{0.2065} \\
Jin Ji (part) & 820 & 119 & 0.1598 & 0.1345 & \textbf{0.1751} & \textbf{0.1597} \\
\bottomrule
\end{tabular}
}
\caption{Cross-corpus pilot on processed \textit{Zizhi Tongjian} subsets. No target-corpus training is performed.}
\label{tab:zztj_transfer}
\end{table}

As shown in Table~\ref{tab:zztj_transfer}, CTD consistently improves both MRR and R@1 on the two \textit{Zizhi Tongjian} subsets without any target-corpus fine-tuning.
Although this pilot is intentionally lighter than the full ChunQiuTR setup, the trend suggests that the temporal-consistency bias learned on ChunQiuTR transfers beyond the source corpus and continues to help distinguish chrono-near but temporally mismatched evidence.

		\subsection{Analysis}
		
		\subsubsection{Impact of Query Type}
		
		\begin{table}[t]
			\centering
			\small
			\setlength{\tabcolsep}{4pt}
			\renewcommand{\arraystretch}{1.15}
			\resizebox{\columnwidth}{!}{%
				\begin{tabular}{l cc cc}
					\toprule
					\textbf{Method} &
					\multicolumn{2}{c}{\textbf{Single-month (\(|Q|=1\))}} &
					\multicolumn{2}{c}{\textbf{Multi-month (\(|Q|>1\))}} \\
					\cmidrule(lr){2-3}\cmidrule(lr){4-5}
					& \textbf{R@1} & \textbf{MRR@10} & \textbf{R@1} & \textbf{MRR@10} \\
					\midrule
					BM25
					& 0.397 & 0.413
					& 0.396 \down{-0.001} & 0.499 \up{+0.086} \\
					
					ColBERT-LFM2$_{(\text{ZS})}$
					& 0.298 & 0.312
					& 0.386 \up{+0.088} & 0.490 \up{+0.178} \\
					
					mE5-Large$_{(\text{ZS})}$
					& 0.259 & 0.279
					& 0.337 \up{+0.078} & 0.422 \up{+0.143} \\
					
					BERT-base$_{(\text{FT})}$
					& 0.497 & 0.516
					& 0.525 \up{+0.027} & 0.621 \up{+0.106} \\
					
					\textbf{\model$_{\text{BERT-base}}$ (Ours)}
					& 0.509 & 0.530
					& 0.685 \up{+0.176} & 0.744 \up{+0.214} \\
					
					Qwen3-Embed-0.6B$_{(\text{ZS})}$
					& 0.353 & 0.385
					& 0.317 \down{-0.036} & 0.415 \up{+0.031} \\
					
					Qwen3-Embed-0.6B$_{(\text{FT})}$
					& 0.481 & 0.495
					& 0.711 \up{+0.230} & 0.757 \up{+0.262} \\
					
					\textbf{\model$_{\text{Qwen3-Embed-0.6B}}$ (Ours)}
					& 0.491 & 0.513
					& 0.733 \up{+0.242} & 0.767 \up{+0.255} \\
					\bottomrule
				\end{tabular}%
			}
			\caption{Impact of query span on retrieval performance on the test set. We compare single-month (\(|Q|=1\)) and multi-month (\(|Q|>1\)) queries; for multi-month queries, the numbers in parentheses give absolute changes relative to single-month queries.}
			\label{tab:impact_query_type}
		\end{table}

        Table~\ref{tab:impact_query_type} compares single-month (\(|Q|=1\)) and multi-month (\(|Q|>1\)) queries. Across most methods, multi-month queries substantially boost MRR@10 and give small gains or no change in R@1, reflecting the fact that it is easier to hit any correct month within a span than a single target month. Our CTD models achieve the best performance in both regimes, with especially large gains on multi-month queries for the BERT backbone (roughly +0.16 R@1) and consistent improvements for Qwen3-Embed, indicating better temporal ordering under chrono-near confounds.

		\subsubsection{Qualitative Examples}
		
		\begin{figure}[t]
			\centering
			\includegraphics[width=\linewidth]{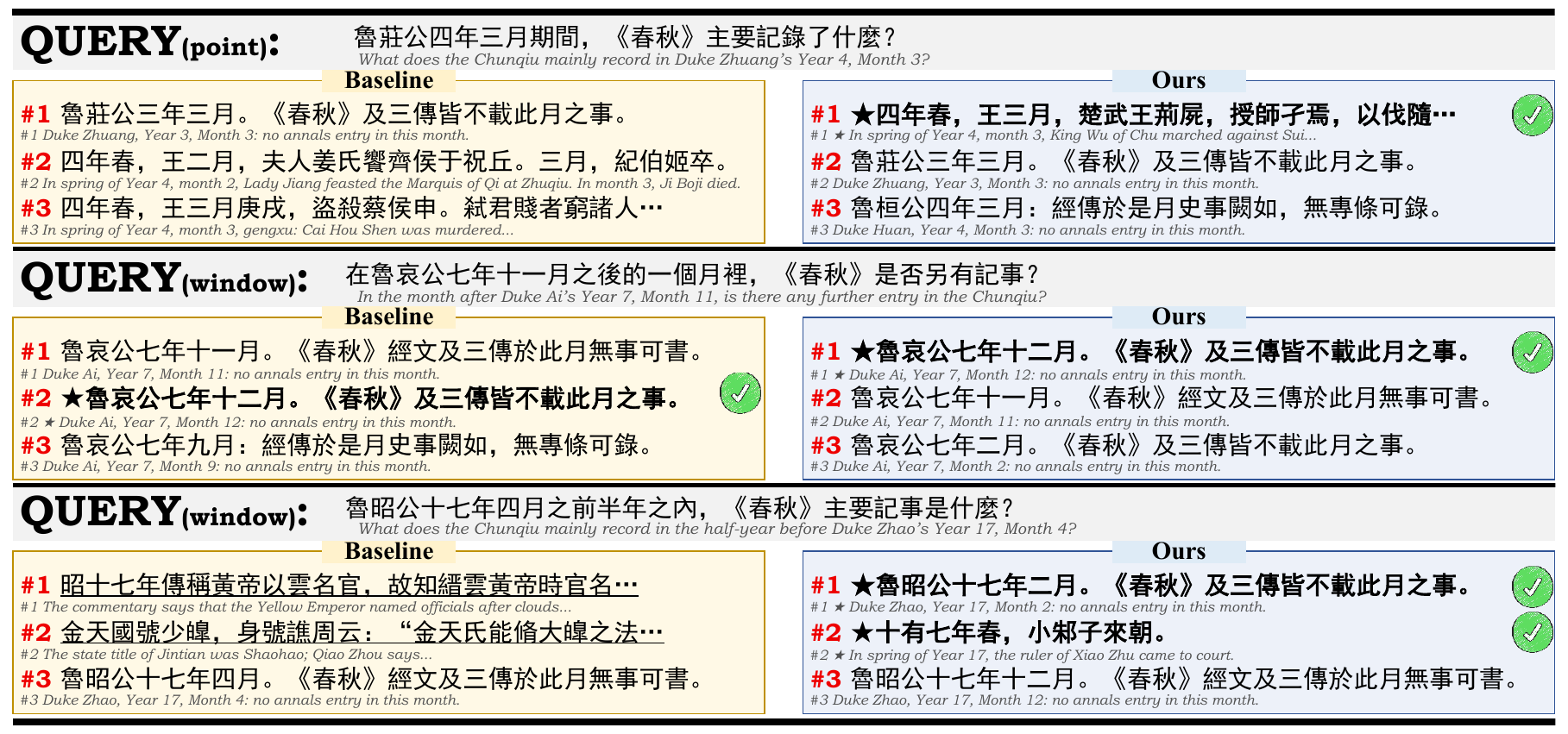}
			\caption{Visualization of Qualitative Examples.}
			\label{fig:main_visualization_demo}
            \vspace{-3mm}
		\end{figure}
		
		Fig.~\ref{fig:main_visualization_demo} illustrates the contrast between a BERT-based baseline and our time-aware retriever via two queries. 
        For a point query, the baseline is distracted by frequent \texttt{no\_event} templates and events from neighboring months, while our model correctly locates the target chronicle entry. 
        For a broader window query, the baseline's results are dispersed across later exegetic discussions, whereas our model concentrates probability on the correct local window, retrieving both the pertinent event and an explicit \texttt{no\_event} record. 
        Overall, these examples show that our retriever couples temporal reasoning with semantic matching beyond surface cues; moreover, temporal errors are often \emph{confident} rather than uncertain, motivating retrieval-time temporal constraints over downstream generation fixes (see Appendix~\ref{sec:app_failure_cases} and Appendix~\ref{sec:app_rag_demo}).

		\subsubsection{Ablation study}
		
		\begin{table}[t]
			\centering
			\small
			\setlength{\tabcolsep}{3pt}
            \resizebox{\linewidth}{!}{%
			\begin{tabular}{l ccc cc}
				\toprule
				\textbf{Variant} &
				$\mathcal{L}_{\text{multi}}$ & $b^{\text{time}}_{ij}$ & $\mathbf{c}_x$ &
				\textbf{R@1} & \textbf{MRR@10} \\
				\midrule
				FT baseline
				& -- & -- & -- &
				\shortstack{0.5771  {\scriptsize ----}} &
				\shortstack{0.6044  {\scriptsize ----}} \\
				
				+ $\mathcal{L}_{\text{multi}}$
				& $\checkmark$ & -- & -- &
				\shortstack{0.5820  {\scriptsize \up{+0.0049}}} &
				\shortstack{0.6107  {\scriptsize \up{+0.0063}}} \\
				
				+ Bias
				& $\checkmark$ & $\checkmark$ & -- &
				\shortstack{0.5898  {\scriptsize \up{+0.0127}}} &
				\shortstack{0.6135  {\scriptsize \up{+0.0091}}} \\
				
				+ Ctx
				& $\checkmark$ & -- & $\checkmark$ &
				\shortstack{0.5850  {\scriptsize \up{+0.0079}}} &
				\shortstack{0.6134  {\scriptsize \up{+0.0090}}} \\
				
				\textbf{Full (Ours)}
				& $\checkmark$ & $\checkmark$ & $\checkmark$ &
				\shortstack{0.5923  {\scriptsize \up{+0.0152}}} &
				\shortstack{0.6194  {\scriptsize \up{+0.0150}}} \\
				\bottomrule
			\end{tabular}%
            }
			\caption{
				Ablation study on the test set under the same evaluation protocol as Table~\ref{tab:main_results}.
				For each metric, the second line reports the change relative to the FT baseline.
			}
			\label{tab:ablation}
            \vspace{-4mm}
		\end{table}

        We ablate three components: the retrieval objective $\mathcal{L}_{\text{multi}}$, the relative-time logit bias $b^{\text{time}}_{ij}$, and the soft absolute temporal context $\mathbf{c}_x$.
        Starting from the FT baseline, adding $\mathcal{L}_{\text{multi}}$ leads to a modest but consistent improvement. This indicates that explicit retrieval supervision enhances time-key discrimination beyond standard fine-tuning.
        Adding either temporal signal further improves performance. The logit bias yields a larger gain in R@1, suggesting that it effectively reshapes in-batch matching toward chronologically plausible candidates. In contrast, injecting temporal context achieves comparable improvements in MRR, indicating better overall ranking quality.
        Combining the bias and the context produces the best results, with additive gains over each individual component. This supports their complementary roles: the bias calibrates pairwise similarities, while the context enriches representations with absolute-time distributional cues.

		\section{Conclusion}
        We presented \textbf{ChunQiuTR}, a time-keyed temporal retrieval dataset built on the \textit{Spring and Autumn Annals} and its commentarial tradition. ChunQiuTR operationalizes a non-Gregorian, reign-based month timeline (gong--year--month) and evaluates retrieval under realistic historical confounders—lexical-near same-key materials, adjacent-month near-misses, and explicit \texttt{no\_event} months—making temporal fidelity a first-class requirement beyond topical relevance. We further proposed \textbf{CTD}, a calendrically time-aware dual-encoder that augments semantic matching with absolute context injection and relative offset biasing. Against strong semantic dual-encoder baselines, CTD consistently improves retrieval quality and reduces chrono-near confusions that can mislead evidence-grounded systems.
		
		\section*{Limitations}
        ChunQiuTR is constructed from the \textit{Chunqiu} annals and its major commentaries, and uses a reign-based month-level time key.
        This narrow scope limits the generality of our findings: retrieval behaviors and error patterns may differ in other pre-modern corpora with different calendrical conventions, narrative styles, and editorial traditions, and our results do not directly imply the same gains under those settings. Extending the same construction procedure to other dynastic corpora would require additional source-specific normalization and time alignment, and we leave such expansions to future work.

        Moreover, a month-level timeline cannot represent finer-grained temporal relations, and remaining errors indicate that (i) near-duplicate records in neighboring months and (ii) genuinely ambiguous historiographical cases are still challenging even with temporal supervision.
        Future work includes extending the benchmark to broader historical corpora and finer temporal granularity, incorporating stronger reranking or evidence-checking for borderline confusions, and evaluating downstream impacts in end-to-end RAG pipelines. 

        \section*{Acknowledgments}

This work was supported in part by the National Natural Science Foundation of China (NSFC) under Grant 62276283 and 62372281, in part by the China Meteorological Administration's Science and Technology Project under Grant CMAJBGS202517, in part by Guangdong-Hong Kong-Macao Greater Bay Area Meteorological Technology Collaborative Research Project under Grant GHMA2024Z04, in part by Fundamental Research Funds for the Central Universities, Sun Yat-sen University under Grant 23hytd006 and 23hytd006-2,  in part by Guangdong Provincial High-Level Young Talent Program under Grant RL2024-151-2-11, in part by the Key Development Project of the Artificial Intelligence Institute, Sun Yat-sen University, and in part by The Major Key Project of PCL (Grant No. PCL2025A17).


        \section*{Ethical Considerations}
        Our data are derived from pre-modern Chinese historical texts (the \textit{Chunqiu} annals and commentarial tradition) and do not contain information about living individuals.
        We use publicly available digital editions and licensed scholarly resources; when redistribution is permitted, we release processed splits and annotations, otherwise we provide code and metadata sufficient to reproduce the benchmark from legitimate sources.
        
        These chronicles reflect historical norms and power structures (e.g., elite-centric perspectives and occasional descriptions of conflict or punishment).
        We preserve such content for historical and linguistic research rather than endorsement.
        We caution against deploying models trained on this benchmark in high-stakes decision-making, and recommend that any downstream RAG or narrative generation make the historical context explicit and include appropriate uncertainty signaling.
        
        We follow the ACL Code of Ethics and complete the Responsible NLP checklist to the best of our knowledge.

		
		\bibliography{custom}

@article{Robertson2009BM25,
author = {Robertson, Stephen and Zaragoza, Hugo},
title = {The Probabilistic Relevance Framework: BM25 and Beyond},
year = {2009},
issue_date = {April 2009},
publisher = {Now Publishers Inc.},
address = {Hanover, MA, USA},
volume = {3},
number = {4},
issn = {1554-0669},
url = {https://doi.org/10.1561/1500000019},
doi = {10.1561/1500000019},
journal = {Found. Trends Inf. Retr.},
month = apr,
pages = {333–389},
numpages = {57}
}

@misc{geng2025competitivesearchrelevanceinferencefree,
      title={Towards Competitive Search Relevance For Inference-Free Learned Sparse Retrievers}, 
      author={Zhichao Geng and Yiwen Wang and Dongyu Ru and Yang Yang},
      year={2025},
      eprint={2411.04403},
      archivePrefix={arXiv},
      primaryClass={cs.IR},
      url={https://arxiv.org/abs/2411.04403}, 
}

@inproceedings{Shen2025sigirExploring,
author = {Shen, Xinjie and Geng, Zhichao and Yang, Yang},
title = {Exploring ℓ0 Sparsification for Inference-free Sparse Retrievers},
year = {2025},
isbn = {9798400715921},
publisher = {Association for Computing Machinery},
address = {New York, NY, USA},
url = {https://doi.org/10.1145/3726302.3730192},
doi = {10.1145/3726302.3730192},
booktitle = {Proceedings of the 48th International ACM SIGIR Conference on Research and Development in Information Retrieval},
pages = {2572–2576},
numpages = {5},
location = {Padua, Italy},
series = {SIGIR '25}
}

@inproceedings{xiao-etal-2024-jina,
    title = "{J}ina-{C}ol{BERT}-v2: A General-Purpose Multilingual Late Interaction Retriever",
    author = {Jha, Rohan  and
      Wang, Bo  and
      G{\"u}nther, Michael  and
      Mastrapas, Georgios  and
      Sturua, Saba  and
      Mohr, Isabelle  and
      Koukounas, Andreas  and
      Wang, Mohammad Kalim  and
      Wang, Nan  and
      Xiao, Han},
    editor = {S{\"a}lev{\"a}, Jonne  and
      Owodunni, Abraham},
    booktitle = "Proceedings of the Fourth Workshop on Multilingual Representation Learning (MRL 2024)",
    month = nov,
    year = "2024",
    address = "Miami, Florida, USA",
    publisher = "Association for Computational Linguistics",
    url = "https://aclanthology.org/2024.mrl-1.11/",
    doi = "10.18653/v1/2024.mrl-1.11",
    pages = "159--166",
}

@misc{amini2025lfm2technicalreport,
      title={LFM2 Technical Report}, 
      author={Liquid AI Team},
      year={2025},
      eprint={2511.23404},
      archivePrefix={arXiv},
      primaryClass={cs.LG},
      url={https://arxiv.org/abs/2511.23404}, 
}

@inproceedings{ni-etal-2022-large-GTR,
    title = "Large Dual Encoders Are Generalizable Retrievers",
    author = "Ni, Jianmo  and
      Qu, Chen  and
      Lu, Jing  and
      Dai, Zhuyun  and
      Hernandez Abrego, Gustavo  and
      Ma, Ji  and
      Zhao, Vincent  and
      Luan, Yi  and
      Hall, Keith  and
      Chang, Ming-Wei  and
      Yang, Yinfei",
    editor = "Goldberg, Yoav  and
      Kozareva, Zornitsa  and
      Zhang, Yue",
    booktitle = "Proceedings of the 2022 Conference on Empirical Methods in Natural Language Processing",
    month = dec,
    year = "2022",
    address = "Abu Dhabi, United Arab Emirates",
    publisher = "Association for Computational Linguistics",
    url = "https://aclanthology.org/2022.emnlp-main.669/",
    doi = "10.18653/v1/2022.emnlp-main.669",
    pages = "9844--9855",
}

@inproceedings{ni-etal-2022-sentence-t5,
    title = "Sentence-T5: Scalable Sentence Encoders from Pre-trained Text-to-Text Models",
    author = "Ni, Jianmo  and
      Hernandez Abrego, Gustavo  and
      Constant, Noah  and
      Ma, Ji  and
      Hall, Keith  and
      Cer, Daniel  and
      Yang, Yinfei",
    editor = "Muresan, Smaranda  and
      Nakov, Preslav  and
      Villavicencio, Aline",
    booktitle = "Findings of the Association for Computational Linguistics: ACL 2022",
    month = may,
    year = "2022",
    address = "Dublin, Ireland",
    publisher = "Association for Computational Linguistics",
    url = "https://aclanthology.org/2022.findings-acl.146/",
    doi = "10.18653/v1/2022.findings-acl.146",
    pages = "1864--1874",
}

@misc{wang2024multilinguale5textembeddings,
      title={Multilingual E5 Text Embeddings: A Technical Report}, 
      author={Liang Wang and Nan Yang and Xiaolong Huang and Linjun Yang and Rangan Majumder and Furu Wei},
      year={2024},
      eprint={2402.05672},
      archivePrefix={arXiv},
      primaryClass={cs.CL},
      url={https://arxiv.org/abs/2402.05672}, 
}

@inproceedings{wang-etal-2024-improving-text,
    title = "Improving Text Embeddings with Large Language Models",
    author = "Wang, Liang  and
      Yang, Nan  and
      Huang, Xiaolong  and
      Yang, Linjun  and
      Majumder, Rangan  and
      Wei, Furu",
    editor = "Ku, Lun-Wei  and
      Martins, Andre  and
      Srikumar, Vivek",
    booktitle = "Proceedings of the 62nd Annual Meeting of the Association for Computational Linguistics (Volume 1: Long Papers)",
    month = aug,
    year = "2024",
    address = "Bangkok, Thailand",
    publisher = "Association for Computational Linguistics",
    url = "https://aclanthology.org/2024.acl-long.642/",
    doi = "10.18653/v1/2024.acl-long.642",
    pages = "11897--11916",
}

@misc{li2023generaltextembeddingsmultistage,
      title={Towards General Text Embeddings with Multi-stage Contrastive Learning}, 
      author={Zehan Li and Xin Zhang and Yanzhao Zhang and Dingkun Long and Pengjun Xie and Meishan Zhang},
      year={2023},
      eprint={2308.03281},
      archivePrefix={arXiv},
      primaryClass={cs.CL},
      url={https://arxiv.org/abs/2308.03281}, 
}

@inproceedings{chen-etal-2024-m3,
    title = "{M}3-Embedding: Multi-Linguality, Multi-Functionality, Multi-Granularity Text Embeddings Through Self-Knowledge Distillation",
    author = "Chen, Jianlyu  and
      Xiao, Shitao  and
      Zhang, Peitian  and
      Luo, Kun  and
      Lian, Defu  and
      Liu, Zheng",
    editor = "Ku, Lun-Wei  and
      Martins, Andre  and
      Srikumar, Vivek",
    booktitle = "Findings of the Association for Computational Linguistics: ACL 2024",
    month = aug,
    year = "2024",
    address = "Bangkok, Thailand",
    publisher = "Association for Computational Linguistics",
    url = "https://aclanthology.org/2024.findings-acl.137/",
    doi = "10.18653/v1/2024.findings-acl.137",
    pages = "2318--2335",
}

@misc{bge_embedding,
      title={C-Pack: Packaged Resources To Advance General Chinese Embedding}, 
      author={Shitao Xiao and Zheng Liu and Peitian Zhang and Niklas Muennighoff},
      year={2023},
      eprint={2309.07597},
      archivePrefix={arXiv},
      primaryClass={cs.CL}
}

@inproceedings{devlin-etal-2019-bert,
    title = "{BERT}: Pre-training of Deep Bidirectional Transformers for Language Understanding",
    author = "Devlin, Jacob  and
      Chang, Ming-Wei  and
      Lee, Kenton  and
      Toutanova, Kristina",
    editor = "Burstein, Jill  and
      Doran, Christy  and
      Solorio, Thamar",
    booktitle = "Proceedings of the 2019 Conference of the North {A}merican Chapter of the Association for Computational Linguistics: Human Language Technologies, Volume 1 (Long and Short Papers)",
    month = jun,
    year = "2019",
    address = "Minneapolis, Minnesota",
    publisher = "Association for Computational Linguistics",
    url = "https://aclanthology.org/N19-1423/",
    doi = "10.18653/v1/N19-1423",
    pages = "4171--4186",
}

@inproceedings{kang-etal-2025-pqr,
    title = "{PQR}: Improving Dense Retrieval via Potential Query Modeling",
    author = "Kang, Junfeng  and
      Li, Rui  and
      Liu, Qi  and
      Chen, Yanjiang  and
      Zhang, Zheng  and
      Jiang, Junzhe  and
      Yu, Heng  and
      Su, Yu",
    editor = "Che, Wanxiang  and
      Nabende, Joyce  and
      Shutova, Ekaterina  and
      Pilehvar, Mohammad Taher",
    booktitle = "Proceedings of the 63rd Annual Meeting of the Association for Computational Linguistics (Volume 1: Long Papers)",
    month = jul,
    year = "2025",
    address = "Vienna, Austria",
    publisher = "Association for Computational Linguistics",
    url = "https://aclanthology.org/2025.acl-long.660/",
    doi = "10.18653/v1/2025.acl-long.660",
    pages = "13455--13469",
    ISBN = "979-8-89176-251-0",
}

@misc{zhang2025qwen3embeddingadvancingtext,
      title={Qwen3 Embedding: Advancing Text Embedding and Reranking Through Foundation Models}, 
      author={Yanzhao Zhang and Mingxin Li and Dingkun Long and Xin Zhang and Huan Lin and Baosong Yang and Pengjun Xie and An Yang and Dayiheng Liu and Junyang Lin and Fei Huang and Jingren Zhou},
      year={2025},
      eprint={2506.05176},
      archivePrefix={arXiv},
      primaryClass={cs.CL},
      url={https://arxiv.org/abs/2506.05176}, 
}

@inproceedings{Formal2021SPLADE,
author = {Formal, Thibault and Piwowarski, Benjamin and Clinchant, St\'{e}phane},
title = {SPLADE: Sparse Lexical and Expansion Model for First Stage Ranking},
year = {2021},
isbn = {9781450380379},
publisher = {Association for Computing Machinery},
address = {New York, NY, USA},
url = {https://doi.org/10.1145/3404835.3463098},
doi = {10.1145/3404835.3463098},
booktitle = {Proceedings of the 44th International ACM SIGIR Conference on Research and Development in Information Retrieval},
pages = {2288–2292},
numpages = {5},
location = {Virtual Event, Canada},
series = {SIGIR '21}
}

@inproceedings{Khattab2020ColBERT,
author = {Khattab, Omar and Zaharia, Matei},
title = {ColBERT: Efficient and Effective Passage Search via Contextualized Late Interaction over BERT},
year = {2020},
isbn = {9781450380164},
publisher = {Association for Computing Machinery},
address = {New York, NY, USA},
url = {https://doi.org/10.1145/3397271.3401075},
doi = {10.1145/3397271.3401075},
booktitle = {Proceedings of the 43rd International ACM SIGIR Conference on Research and Development in Information Retrieval},
pages = {39–48},
numpages = {10},
location = {Virtual Event, China},
series = {SIGIR '20}
}

@inproceedings{karpukhin-etal-2020-dense,
    title = "Dense Passage Retrieval for Open-Domain Question Answering",
    author = "Karpukhin, Vladimir  and
      Oguz, Barlas  and
      Min, Sewon  and
      Lewis, Patrick  and
      Wu, Ledell  and
      Edunov, Sergey  and
      Chen, Danqi  and
      Yih, Wen-tau",
    editor = "Webber, Bonnie  and
      Cohn, Trevor  and
      He, Yulan  and
      Liu, Yang",
    booktitle = "Proceedings of the 2020 Conference on Empirical Methods in Natural Language Processing (EMNLP)",
    month = nov,
    year = "2020",
    address = "Online",
    publisher = "Association for Computational Linguistics",
    url = "https://aclanthology.org/2020.emnlp-main.550/",
    doi = "10.18653/v1/2020.emnlp-main.550",
    pages = "6769--6781",
}

@inproceedings{
xiong2021approximate,
title={Approximate Nearest Neighbor Negative Contrastive Learning for Dense Text Retrieval},
author={Lee Xiong and Chenyan Xiong and Ye Li and Kwok-Fung Tang and Jialin Liu and Paul N. Bennett and Junaid Ahmed and Arnold Overwijk},
booktitle={International Conference on Learning Representations},
year={2021},
url={https://openreview.net/forum?id=zeFrfgyZln}
}

@inproceedings{lei-etal-2023-unsupervised,
    title = "Unsupervised Dense Retrieval with Relevance-Aware Contrastive Pre-Training",
    author = "Lei, Yibin  and
      Ding, Liang  and
      Cao, Yu  and
      Zan, Changtong  and
      Yates, Andrew  and
      Tao, Dacheng",
    editor = "Rogers, Anna  and
      Boyd-Graber, Jordan  and
      Okazaki, Naoaki",
    booktitle = "Findings of the Association for Computational Linguistics: ACL 2023",
    month = jul,
    year = "2023",
    address = "Toronto, Canada",
    publisher = "Association for Computational Linguistics",
    url = "https://aclanthology.org/2023.findings-acl.695/",
    doi = "10.18653/v1/2023.findings-acl.695",
    pages = "10932--10940",
}

@inproceedings{li2003timebasedlm,
author = {Li, Xiaoyan and Croft, W. Bruce},
title = {Time-based language models},
year = {2003},
isbn = {1581137230},
publisher = {Association for Computing Machinery},
address = {New York, NY, USA},
url = {https://doi.org/10.1145/956863.956951},
doi = {10.1145/956863.956951},
booktitle = {Proceedings of the Twelfth International Conference on Information and Knowledge Management},
pages = {469–475},
numpages = {7},
location = {New Orleans, LA, USA},
series = {CIKM '03}
}

@misc{piryani2025itshightimesurvey,
      title={It's High Time: A Survey of Temporal Question Answering}, 
      author={Bhawna Piryani and Abdelrahman Abdallah and Jamshid Mozafari and Avishek Anand and Adam Jatowt},
      year={2025},
      eprint={2505.20243},
      archivePrefix={arXiv},
      primaryClass={cs.CL},
      url={https://arxiv.org/abs/2505.20243}, 
}

@inproceedings{Rajapakse2023densePassage,
author = {Rajapakse, Thilina C.},
title = {Dense Passage Retrieval: Architectures and Augmentation Methods},
year = {2023},
isbn = {9781450394086},
publisher = {Association for Computing Machinery},
address = {New York, NY, USA},
url = {https://doi.org/10.1145/3539618.3591796},
doi = {10.1145/3539618.3591796},
booktitle = {Proceedings of the 46th International ACM SIGIR Conference on Research and Development in Information Retrieval},
pages = {3494},
numpages = {1},
location = {Taipei, Taiwan},
series = {SIGIR '23}
}

@inproceedings{han-etal-2025-temporal,
    title = "Temporal Information Retrieval via Time-Specifier Model Merging",
    author = "Han, SeungYoon  and
      Hwang, Taeho  and
      Cho, Sukmin  and
      Jeong, Soyeong  and
      Song, Hoyun  and
      Lee, Huije  and
      Park, Jong C.",
    editor = "Zhang, Yuji  and
      Chen, Canyu  and
      Li, Sha  and
      Geva, Mor  and
      Han, Chi  and
      Wang, Xiaozhi  and
      Feng, Shangbin  and
      Gao, Silin  and
      Augenstein, Isabelle  and
      Bansal, Mohit  and
      Li, Manling  and
      Ji, Heng",
    booktitle = "Proceedings of the 3rd Workshop on Towards Knowledgeable Foundation Models (KnowFM)",
    month = aug,
    year = "2025",
    address = "Vienna, Austria",
    publisher = "Association for Computational Linguistics",
    url = "https://aclanthology.org/2025.knowllm-1.1/",
    doi = "10.18653/v1/2025.knowllm-1.1",
    pages = "1--13",
    ISBN = "979-8-89176-283-1",
}

@inproceedings{zhang2025mrag,
  title     = {MRAG: A Modular Retrieval Framework for Time-Sensitive Question Answering},
  author    = {Zhang, Siyue and Xue, Yuxiang and Zhang, Yiming and Wu, Xiaobao and Luu, Anh Tuan and Zhao, Chen},
  booktitle = {Findings of the Association for Computational Linguistics: EMNLP 2025},
  year      = {2025},
  pages     = {3080--3118}
}

@InProceedings{pmlr-v202-yeche23a,
  title = 	 {Temporal Label Smoothing for Early Event Prediction},
  author =       {Y\`{e}che, Hugo and Pace, Aliz\'{e}e and Ratsch, Gunnar and Kuznetsova, Rita},
  booktitle = 	 {Proceedings of the 40th International Conference on Machine Learning},
  pages = 	 {39913--39938},
  year = 	 {2023},
  editor = 	 {Krause, Andreas and Brunskill, Emma and Cho, Kyunghyun and Engelhardt, Barbara and Sabato, Sivan and Scarlett, Jonathan},
  volume = 	 {202},
  series = 	 {Proceedings of Machine Learning Research},
  month = 	 {23--29 Jul},
  publisher =    {PMLR},
  url = 	 {https://proceedings.mlr.press/v202/yeche23a.html},
}

@inproceedings{wang2023BiTimeBERT,
author = {Wang, Jiexin and Jatowt, Adam and Yoshikawa, Masatoshi and Cai, Yi},
title = {BiTimeBERT: Extending Pre-Trained Language Representations with Bi-Temporal Information},
year = {2023},
isbn = {9781450394086},
publisher = {Association for Computing Machinery},
address = {New York, NY, USA},
url = {https://doi.org/10.1145/3539618.3591686},
doi = {10.1145/3539618.3591686},
booktitle = {Proceedings of the 46th International ACM SIGIR Conference on Research and Development in Information Retrieval},
pages = {812–821},
numpages = {10},
location = {Taipei, Taiwan},
series = {SIGIR '23}
}

@article{dhingra-etal-2022-time,
    title = "Time-Aware Language Models as Temporal Knowledge Bases",
    author = "Dhingra, Bhuwan  and
      Cole, Jeremy R.  and
      Eisenschlos, Julian Martin  and
      Gillick, Daniel  and
      Eisenstein, Jacob  and
      Cohen, William W.",
    editor = "Roark, Brian  and
      Nenkova, Ani",
    journal = "Transactions of the Association for Computational Linguistics",
    volume = "10",
    year = "2022",
    address = "Cambridge, MA",
    publisher = "MIT Press",
    url = "https://aclanthology.org/2022.tacl-1.15/",
    doi = "10.1162/tacl_a_00459",
    pages = "257--273",
}

@article{gao2023retrieval,
  title={Retrieval-augmented generation for large language models: A survey},
  author={Gao, Yunfan and Xiong, Yun and Gao, Xinyu and Jia, Kangxiang and Pan, Jinliu and Bi, Yuxi and Dai, Yixin and Sun, Jiawei and Wang, Haofen and Wang, Haofen},
  journal={arXiv preprint arXiv:2312.10997},
  volume={2},
  number={1},
  year={2023}
}

@inproceedings{Lewis2020Retrieval,
 author = {Lewis, Patrick and Perez, Ethan and Piktus, Aleksandra and Petroni, Fabio and Karpukhin, Vladimir and Goyal, Naman and K\"{u}ttler, Heinrich and Lewis, Mike and Yih, Wen-tau and Rockt\"{a}schel, Tim and Riedel, Sebastian and Kiela, Douwe},
 booktitle = {Advances in Neural Information Processing Systems},
 editor = {H. Larochelle and M. Ranzato and R. Hadsell and M.F. Balcan and H. Lin},
 pages = {9459--9474},
 publisher = {Curran Associates, Inc.},
 title = {Retrieval-Augmented Generation for Knowledge-Intensive NLP Tasks},
 url = {https://proceedings.neurips.cc/paper_files/paper/2020/file/6b493230205f780e1bc26945df7481e5-Paper.pdf},
 volume = {33},
 year = {2020}
}

@article{menick2022teaching,
  title={Teaching language models to support answers with verified quotes},
  author={Menick, Jacob and Trebacz, Maja and Mikulik, Vladimir and Aslanides, John and Song, Francis and Chadwick, Martin and Glaese, Mia and Young, Susannah and Campbell-Gillingham, Lucy and Irving, Geoffrey and others},
  journal={arXiv preprint arXiv:2203.11147},
  year={2022}
}

@inproceedings{cao-etal-2024-tonggu,
    title = "{T}ong{G}u: Mastering Classical {C}hinese Understanding with Knowledge-Grounded Large Language Models",
    author = "Cao, Jiahuan  and
      Peng, Dezhi  and
      Zhang, Peirong  and
      Shi, Yongxin  and
      Liu, Yang  and
      Ding, Kai  and
      Jin, Lianwen",
    editor = "Al-Onaizan, Yaser  and
      Bansal, Mohit  and
      Chen, Yun-Nung",
    booktitle = "Findings of the Association for Computational Linguistics: EMNLP 2024",
    month = nov,
    year = "2024",
    address = "Miami, Florida, USA",
    publisher = "Association for Computational Linguistics",
    url = "https://aclanthology.org/2024.findings-emnlp.243/",
    doi = "10.18653/v1/2024.findings-emnlp.243",
    pages = "4196--4210"
}

@inproceedings{zhang-etal-2024-philogpt,
    title = "{P}hilo{GPT}: A Philology-Oriented Large Language Model for {A}ncient {C}hinese Manuscripts with Dunhuang as Case Study",
    author = "Zhang, Yuqing  and
      He, Baoyi  and
      Chen, Yihan  and
      Li, Hangqi  and
      Yue, Han  and
      Zhang, Shengyu  and
      Dou, Huaiyong  and
      Yan, Junchi  and
      Liu, Zemin  and
      Zhang, Yongquan  and
      Wu, Fei",
    editor = "Al-Onaizan, Yaser  and
      Bansal, Mohit  and
      Chen, Yun-Nung",
    booktitle = "Proceedings of the 2024 Conference on Empirical Methods in Natural Language Processing",
    month = nov,
    year = "2024",
    address = "Miami, Florida, USA",
    publisher = "Association for Computational Linguistics",
    url = "https://aclanthology.org/2024.emnlp-main.163/",
    doi = "10.18653/v1/2024.emnlp-main.163",
    pages = "2784--2801",
}

@inproceedings{liu-etal-2025-large,
    title = "Large-Scale Corpus Construction and Retrieval-Augmented Generation for {A}ncient {C}hinese Poetry: New Method and Data Insights",
    author = "Liu, Yang  and
      Lan, Lan  and
      Cao, Jiahuan  and
      Cheng, Hiuyi  and
      Ding, Kai  and
      Jin, Lianwen",
    editor = "Chiruzzo, Luis  and
      Ritter, Alan  and
      Wang, Lu",
    booktitle = "Findings of the Association for Computational Linguistics: NAACL 2025",
    month = apr,
    year = "2025",
    address = "Albuquerque, New Mexico",
    publisher = "Association for Computational Linguistics",
    url = "https://aclanthology.org/2025.findings-naacl.46/",
    doi = "10.18653/v1/2025.findings-naacl.46",
    pages = "779--817",
    ISBN = "979-8-89176-195-7",
}

@inproceedings{chen2021timeqa,
 author = {Chen, Wenhu and Wang, Xinyi and Wang, William Yang and Wang, William Yang},
 booktitle = {Proceedings of the Neural Information Processing Systems Track on Datasets and Benchmarks},
 editor = {J. Vanschoren and S. Yeung},
 pages = {},
 title = {A Dataset for Answering Time-Sensitive Questions},
 url = {https://datasets-benchmarks-proceedings.neurips.cc/paper_files/paper/2021/file/1f0e3dad99908345f7439f8ffabdffc4-Paper-round2.pdf},
 volume = {1},
 year = {2021}
}

@Article{Chen2025tempqa,
author={Chen, Ziyang
and Min, Erxue
and Zhao, Xiang
and Li, Yunxin
and Jia, Xin
and Liao, Jinzhi
and Li, Jichao
and Wang, Shuaiqiang
and Hu, Baotian
and Yin, Dawei},
title={A Question Answering Dataset for Temporal-Sensitive Retrieval-Augmented Generation},
journal={Scientific Data},
year={2025},
month={Nov},
day={21},
volume={12},
number={1},
pages={1855},
issn={2052-4463},
doi={10.1038/s41597-025-06098-y},
url={https://doi.org/10.1038/s41597-025-06098-y}
}

@InProceedings{pmlr-v119-chen20j,
  title = 	 {A Simple Framework for Contrastive Learning of Visual Representations},
  author =       {Chen, Ting and Kornblith, Simon and Norouzi, Mohammad and Hinton, Geoffrey},
  booktitle = 	 {Proceedings of the 37th International Conference on Machine Learning},
  pages = 	 {1597--1607},
  year = 	 {2020},
  editor = 	 {III, Hal Daumé and Singh, Aarti},
  volume = 	 {119},
  series = 	 {Proceedings of Machine Learning Research},
  month = 	 {13--18 Jul},
  publisher =    {PMLR},
  url = 	 {https://proceedings.mlr.press/v119/chen20j.html},
}

@inproceedings{Cao2022XLTime,
    title = "{XLT}ime: A Cross-Lingual Knowledge Transfer Framework for Temporal Expression Extraction",
    author = "Cao, Yuwei  and
      Groves, William  and
      Saha, Tanay Kumar  and
      Tetreault, Joel  and
      Jaimes, Alejandro  and
      Peng, Hao  and
      Yu, Philip",
    editor = "Carpuat, Marine  and
      de Marneffe, Marie-Catherine  and
      Meza Ruiz, Ivan Vladimir",
    booktitle = "Findings of the Association for Computational Linguistics: NAACL 2022",
    month = jul,
    year = "2022",
    address = "Seattle, United States",
    publisher = "Association for Computational Linguistics",
    url = "https://aclanthology.org/2022.findings-naacl.148/",
    doi = "10.18653/v1/2022.findings-naacl.148",
    pages = "1931--1942"
}

@inproceedings{Korchagina2016,
  title={Building a Gold Standard for Temporal Entity Extraction from Medieval German Texts},
  author={Natalia Korchagina},
  year={2016},
  booktitle={"2016 Conference on 
Language Technologies and Digital Humanities"},
  url={https://api.semanticscholar.org/CorpusID:70088125}
}

@article{SanchezDeCastro2025TemporalNormalization,
    author = {Castro, Alejandro Sánchez de and Araujo, Lourdes and Martinez-Romo, Juan},
    title = {A Novel Methodology for Enhancing Cross-language and Domain Adaptability in Temporal Expression Normalization},
    journal = {Computational Linguistics},
    volume = {51},
    number = {4},
    pages = {1303-1335},
    year = {2025},
    month = {12},
    issn = {0891-2017},
    doi = {10.1162/COLI.a.12},
    url = {https://doi.org/10.1162/COLI.a.12},
    eprint = {https://direct.mit.edu/coli/article-pdf/51/4/1303/2523155/coli.a.12.pdf},
}

@inproceedings{Su2025TemporalIEReview,
    title = "Transformer-Based Temporal Information Extraction and Application: A Review",
    author = "Su, Xin  and
      Howard, Phillip  and
      Bethard, Steven",
    editor = "Christodoulopoulos, Christos  and
      Chakraborty, Tanmoy  and
      Rose, Carolyn  and
      Peng, Violet",
    booktitle = "Proceedings of the 2025 Conference on Empirical Methods in Natural Language Processing",
    month = nov,
    year = "2025",
    address = "Suzhou, China",
    publisher = "Association for Computational Linguistics",
    url = "https://aclanthology.org/2025.emnlp-main.1467/",
    doi = "10.18653/v1/2025.emnlp-main.1467",
    pages = "28822--28841",
    ISBN = "979-8-89176-332-6"
}

@inproceedings{Graciotti2025KEMHISTO,
    title = "{KE}-{MHISTO}: Towards a Multilingual Historical Knowledge Extraction Benchmark for Addressing the Long-Tail Problem",
    author = "Graciotti, Arianna  and
      Piano, Leonardo  and
      Lazzari, Nicolas  and
      Daga, Enrico  and
      Tripodi, Rocco  and
      Presutti, Valentina  and
      Pompianu, Livio",
    editor = "Che, Wanxiang  and
      Nabende, Joyce  and
      Shutova, Ekaterina  and
      Pilehvar, Mohammad Taher",
    booktitle = "Findings of the Association for Computational Linguistics: ACL 2025",
    month = jul,
    year = "2025",
    address = "Vienna, Austria",
    publisher = "Association for Computational Linguistics",
    url = "https://aclanthology.org/2025.findings-acl.1042/",
    doi = "10.18653/v1/2025.findings-acl.1042",
    pages = "20316--20339",
    ISBN = "979-8-89176-256-5"
}

		\appendix
		
		\section{Details of Dataset}

		\begin{table*}[t]
    \centering
    \footnotesize
    \setlength{\tabcolsep}{3.5pt}
    \renewcommand{\arraystretch}{1.04}
    \begin{tabularx}{\textwidth}{p{0.70\textwidth}p{0.18\textwidth}p{0.07\textwidth}}
        \toprule
        Text (excerpt + short gloss) & Source & Label \\
        \midrule

        \textbf{夏，五月，郑伯克段于鄢。}\\
        \emph{Annals anchor: Zheng subdued Duan at Yan in month 5.}
        & \shortstack[l]{《春秋》\\\textit{Chunqiu}\\经文 / \textit{annals}}
        & event \\
        \midrule

        \textbf{夏，五月，郑伯克段于鄢。克之者何？杀之也。杀之则曷为谓之克？大郑伯之恶也。段者何？郑伯之弟也。何以不称弟？当国也。}\\
        \emph{Gongyang reads \textit{ke} as killing and amplifies moral blame.}
        & \shortstack[l]{《春秋公羊传》\\\textit{Gongyang zhuan}\\传文 / \textit{zhuan}}
        & event \\
        \midrule

        \textbf{夏，五月，郑伯克段于鄢。克者何？能也。何能也？能杀也。何以不言杀？见段之有徒众也。段，郑伯弟也……段失子弟之道矣，贱段而甚郑伯也。}\\
        \emph{Guliang stresses armed revolt, conduct, and intensified censure.}
        & \shortstack[l]{《春秋谷梁传》\\\textit{Guliang zhuan}\\传文 / \textit{zhuan}}
        & event \\
        \midrule

        \textbf{书曰：“郑伯克段于鄢。”段不弟，故不言弟；如二君，故曰克；称郑伯，讥失教也，谓之郑志。不言出奔，难之也。}\\
        \emph{Zuo treats wording itself as a signal of layered blame.}
        & \shortstack[l]{《春秋左传》\\\textit{Zuo zhuan}\\传文 / \textit{zhuan}}
        & event \\
        \midrule

        \textbf{案左氏云段出奔共，而公、谷皆曰杀。据隐十一年传，庄公曰：“寡人有弟不能和协，使糊其口于四方”，则未杀明矣，公、谷之说非是。}\\
        \emph{Gu Donggao rejects the reading that Duan was killed.}
        & \shortstack[l]{《春秋大事表》\\\textit{Chronological}\\顾栋高 / \textit{Qing}}
        & neg \\
        \midrule

        \textbf{不称国讨而言郑伯，讥失教也。段不弟，故不言弟，明郑伯虽失教而段亦凶逆。}\\
        \emph{Du Yu explains how wording encodes both political and personal blame.}
        & \shortstack[l]{《春秋左传注》\\\textit{Zuo zhuan zhu}\\杜预 / \textit{Jin}}
        & neg \\
        \midrule

        \textbf{以“国讨”“得隽曰克”等例，说明称郑伯乃罪君，不称弟乃罪段，兼示兄虽失教而弟为乱首。}\\
        \emph{Kong Yingda systematizes the case through doctrinal categories.}
        & \shortstack[l]{《春秋左传疏》\\\textit{Zuo zhuan shu}\\孔颖达 / \textit{Tang}}
        & neg \\
        \bottomrule
    \end{tabularx}
    \caption{Aligned materials under the reign-based time key ``鲁隐公元年五月'' for ``郑伯克段于鄢''. The annals give the anchor, the three \textit{zhuan} provide aligned expansions, and later sources yield chrono-near non-target paraphrases.}
    \label{tab:yin1-may-ke-duan}
\end{table*}

        \subsection{Annals and Exegetical Layers}
\label{sec:appendix_detail_data}

The \textit{Chunqiu} corpus combines two tightly coupled layers. The annals themselves are extremely terse month-level records, which provide the primary temporal anchors on the Lu-state regnal timeline. By contrast, the three classical \textit{zhuan} (\textit{Zuo}, \textit{Gongyang}, and \textit{Guliang}) expand the same entries into narrative, interpretive, or doctrinal prose. In ChunQiuTR, we treat the annals line as the anchor and the aligned \textit{zhuan} passages as semantically richer event descriptions under the same reign-based time key.

Table~\ref{tab:yin1-may-ke-duan} illustrates this structure with the canonical case ``郑伯克段于鄢'' (\emph{Duke Yin, Year 1, Month 5}). A single annals line is expanded by the three \textit{zhuan} in different ways, while later commentarial and historiographical sources further paraphrase or reinterpret the same event. This layered organization is central to our benchmark design: it provides both aligned event records and naturally occurring chrono-near but non-target passages.

		\subsection{From Parallel Texts to Reign-Based Time Keys}

\subsubsection{Reign-based time keys}
\label{sec:reign-time-key-details}

Our normalized time axis uses month-level keys of the form
\[
\tau=(\text{gong},\text{year},\text{month}),
\]
for example 「鲁隐公元年正月」 or 「鲁桓公元年三月」.  
We scan the annals sequentially while maintaining the current triple \((g,y,m)\). Full reign cues initialize a new key, bare year markers start a new regnal year under the current ruler, month markers update only \(m\), and sentences without new temporal cues inherit the current triple. Table~\ref{tab:reign-time-key} shows representative cases.

\begin{table*}[t]
    \centering
    \footnotesize
    \setlength{\tabcolsep}{4pt}
    \renewcommand{\arraystretch}{1.05}
    \begin{tabular}{p{0.42\linewidth}p{0.30\linewidth}p{0.20\linewidth}}
        \toprule
        Original chronicle snippet & Cue / update & Normalized time key \\
        \midrule
        元年，春，王正月。
        & Initialize a new reign-year (隐公元年), month = 正月
        & 鲁隐公元年正月 \\
        
        三月，公及邾仪父盟于蔑。
        & New month (三月), inherit current reign-year
        & 鲁隐公元年三月 \\
        
        二年，春，公会戎于潜。
        & New year (二年) under the same ruler, reset month to 正月
        & 鲁隐公二年正月 \\
        
        元年，春，王正月，公即位。
        & New ruler detected (桓公), reset year to 元年 and month to 正月
        & 鲁桓公元年正月 \\
        \bottomrule
    \end{tabular}
    \caption{Representative mappings from raw chronicle phrases to normalized reign-based time keys.}
    \label{tab:reign-time-key}
\end{table*}
		
		\subsubsection{Record--time-key alignment}
\label{sec:appendix_record_alignment}

\begin{figure}[tb]
    \includegraphics[width=0.85\columnwidth]{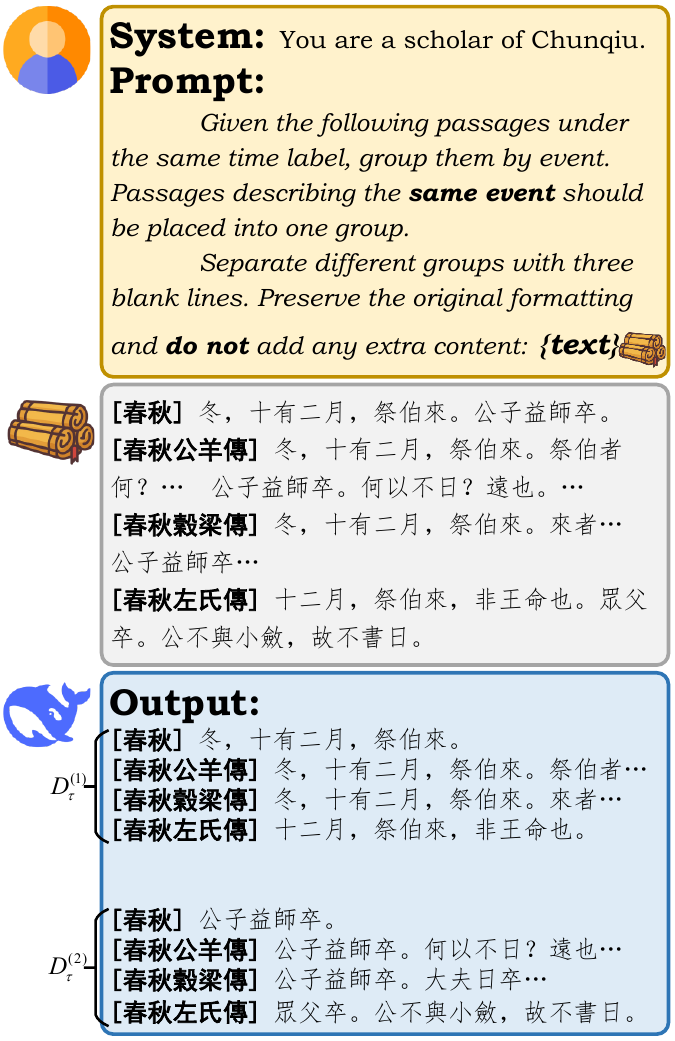}
    \caption{Example of event-level grouping under the reign-based time key 「鲁隐公元年十二月」. The model is prompted to split mixed passages and group aligned commentary snippets by event.}
    \label{fig:appendix_demo_event_alignment}
\end{figure}

We align event-level records under each normalized time key using lightweight LLM suggestions followed by manual verification. This is necessary because a single annals line may compress multiple events, while commentary passages may expand one event across several fragments. Figure~\ref{fig:appendix_demo_event_alignment} shows a representative case under the time key 「鲁隐公元年十二月」.

In this case, the mixed annals line is split into two records, one for 「祭伯来」 and one for 「公子益师卒」, each grouped with its aligned \textit{zhuan} passages. All such suggestions are manually checked before entering the benchmark.

		\subsection{Details of Chrono-near counterfactual negatives}
		\label{sec:appendix_counterfactual}
		
		\subsubsection{Classical sources and perspectives}
		
		Our chrono-near counterfactual negatives draw on several classical works that reorganize or reinterpret \textit{Chunqiu} events from distinct perspectives:
		
		\paragraph{Gu Donggao’s \textit{Chronological Tables} (顾栋高《春秋大事表》).} Gu’s Qing-dynasty compilation systematically re-orders the \textit{Chunqiu} and the three \textit{zhuan} into explicit chronological tables. Each entry typically specifies the state, the reigning gong, the year, and a short prose summary of the event, sometimes highlighting cross-state interactions or disagreements among the Zuo, Gongyang, and Guliang traditions. Compared to the terse annals, these tables provide a more “modernized” timeline and condensed paraphrases of events, which we reuse as temporally grounded, paraphrastic negatives.
		
		\paragraph{Wei Liaoweng’s \textit{Chunqiu Zuozhuan Yaoyi} (魏了翁《春秋左传要义》).} Wei’s Southern Song work focuses on extracting the “essential meanings” of \textit{Zuo zhuan} episodes. His prose often paraphrases the underlying narrative, emphasizes moral and ritual judgments, and occasionally re-groups several \textit{Zuo} passages into a single didactic unit. From our perspective, these are high-level, discursive restatements of the same historical events, written in a style that differs noticeably from the base corpus.
		
		\paragraph{Zuoshu annotations and sub-commentaries (注疏).} In addition, we employ the \textit{Siku Quanshu} edition of \textit{Zuozhuan} annotations, which combines multiple layers: Lu Deming’s \textit{yin yi} (音义), Du Yu’s Jin-dynasty commentary, Kong Yingda’s Tang-dynasty \textit{Chunqiu Zhengyi}, and later Song-dynasty notes such as Lü Zuqian’s \textit{Zuozhuan shuō}. These texts embed glosses, philological notes, and exegetical reformulations around the same events. While they do not always restate the full narrative, they frequently echo key phrases, name important actors, or re-frame the event in ritual or moral terms.
		
		Taken together, these sources provide us with multiple “views” on the same historical episodes: terse annal entries, narrative expansions in the three \textit{zhuan}, tabular re-organizations (Gu Donggao), moral-didactic summaries (Wei Liaoweng), and layered annotations (注疏). By aligning them into the reign-based time-key space defined in the main text, we obtain chrono-near passages that are temporally co-located with our ground truth records \(D_{\tau}\) but often differ in wording, emphasis, or even event granularity.
		
		\subsubsection{LLM-assisted reverse matching and fuzzy alignment}
		
		\begin{figure}[t]
			\centering
			\includegraphics[width=\columnwidth]{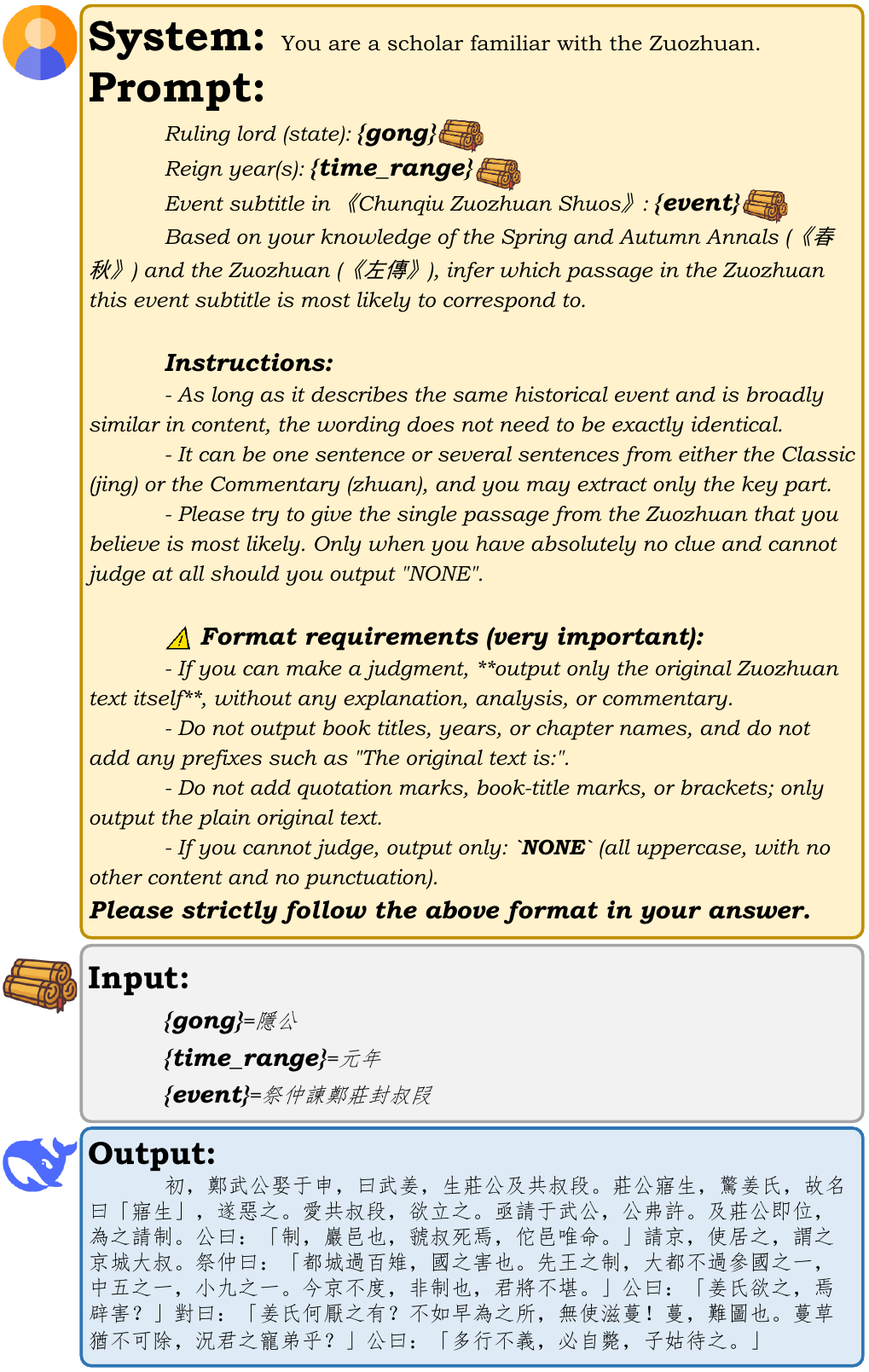}
			\caption{LLM-assisted reverse matching from paraphrastic event titles in Lü Zuqian's \emph{Chunqiu ZuoZhuan Shuos} (《春秋左氏传说》) to \textit{Zuo zhuan} passages. We show the full text-only prompt given to a classical-Chinese LLM (DeepSeek), together with one concrete example: for ruler \textit{Yin} of Lu, year 1, and the subtitle “祭仲谏郑庄封叔段” from \emph{Chunqiu ZuoZhuan Shuos}, the model proposes the most likely \textit{Zuo zhuan} passage. The model is required to output only the original \textit{Zuo} text, or the sentinel token \texttt{NONE} when it cannot decide.}
			\label{fig:appendix_demo_reverse_alignment}
		\end{figure}
		
		For sources like Lü Zuqian's \emph{Chunqiu ZuoZhuan Shuos} (《春秋左氏传说》) and certain annotation layers, the text is often organized as short titled sections (e.g., event summaries or topic headings) rather than direct quotations of the base corpus. To map these paraphrastic units back to concrete passages in the annals and \textit{Zuo zhuan}, we adopt an LLM-assisted ``reverse matching'' strategy, illustrated in Fig.~\ref{fig:appendix_demo_reverse_alignment}.
		
		Given a candidate item with a ruler name, an approximate year range, and a short event title (e.g., from Lü's \emph{Chunqiu ZuoZhuan Shuos}), we query a classical-Chinese LLM (DeepSeek) as a virtual \textit{Zuo zhuan} expert. When the model can make a judgment, it must return only the original \textit{Zuo zhuan} text segment; when it is uncertain, it must output the sentinel token \texttt{NONE}. In all cases, these LLM suggestions are further filtered and manually checked before being accepted into our aligned record set.
		
		Once a plausible \textit{Zuo zhuan} span has been suggested and validated, it can be matched back to the digitized base text with simple fuzzy string matching, which uniquely anchors the passage to its canonical location and the corresponding reign-based time key \(\tau\).

		\subsection{Query Types and Templates}
\label{sec:appendix_queries}

\begin{figure*}[t]
    \includegraphics[width=\linewidth]{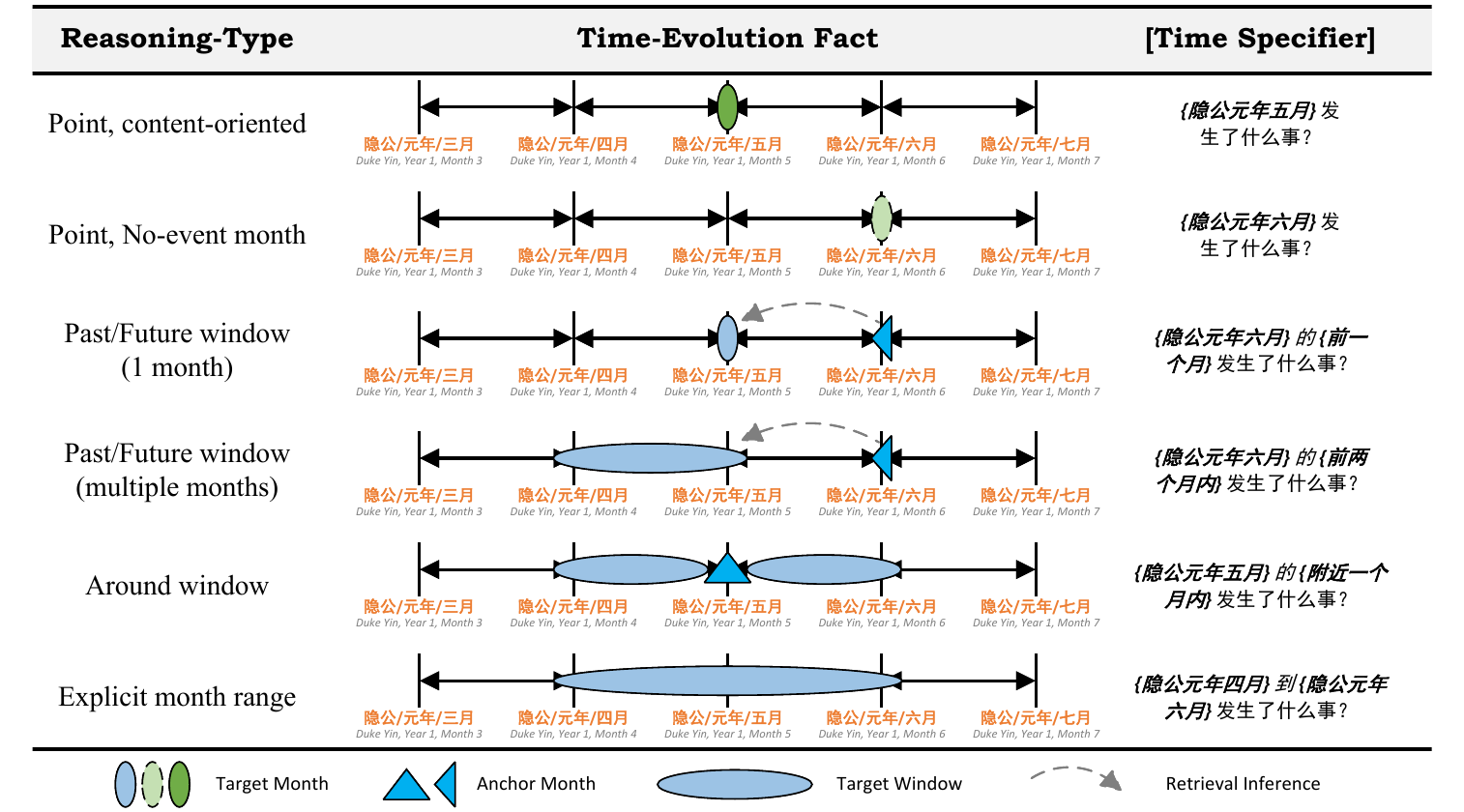}
    \caption{Representative temporal query types in ChunQiuTR, including point queries, past/future windows, around windows, and explicit ranges over reign-based month keys.}
    \label{fig:app_temporal_reasoning_patterns}
\end{figure*}

\begin{table}[t]
    \centering
    \small
    \resizebox{\columnwidth}{!}{%
        \begin{tabular}{ll}
            \toprule
            Template groups (index range) & Type \\
            \midrule
            BASE (1--12) & point, content-oriented \\
            BASE (13--20) & point, existence / no-event \\
            \midrule
            MONTH\_PAST (21--26) & window, past \\
            MONTH\_FUTURE (27--31) & window, future \\
            MONTH\_AROUND (32--36) & window, around \\
            MONTH\_RANGE (37--41) & window, range \\
            \midrule
            YEAR\_CURRENT (42--46) & window, current year \\
            YEAR\_PAST (47--49) & window, previous year \\
            YEAR\_FUTURE (50--52) & window, next year \\
            \bottomrule
        \end{tabular}%
    }
    \caption{Query template groups used in ChunQiuTR.}
    \label{tab:query_template_types}
\end{table}

We instantiate a small set of Traditional-Chinese natural-language templates over normalized reign-based month keys (e.g., ``魯隱公元年二月''). Queries are divided into \emph{point} queries, which target a single month, and \emph{window} queries, which target a span around a reference month. Point queries include both content-oriented and existence-oriented formulations, while window queries cover past, future, around, and explicit-range retrieval. Figure~\ref{fig:app_temporal_reasoning_patterns} illustrates representative temporal interpretations, and Table~\ref{tab:query_template_types} summarizes the template groups used in experiments. Empty months are handled through the same point/window formulations via \texttt{no\_event} placeholders when the target month or span contains no recorded event.

		\subsection{Statistics of Dataset}
		\label{sec:appendix_statistic_month}
		
		\subsubsection{Raw unit-level length statistics (before sentence splitting)}
		
		Before sentence-level segmentation, we collect raw records from the \emph{Chunqiu} annals and the three traditional \emph{zhuan} (positive pool), as well as later exegetical layers such as \emph{Zuoshi zhuanshuo} and \emph{Chunqiu Zhengyi} (negative pool). Table~\ref{tab:raw_length_stats} reports basic character-length statistics over these raw records.

		\begin{table*}[t]
			\centering
			\small
			\resizebox{\textwidth}{!}{%
				\begin{tabular}{llrrrrrr}
					\toprule
					Pool & Source & \# Raw units & Total chars (K) & Avg. len. & Median & Min & Max \\
					\midrule
					Positive & All (annals + three \emph{zhuan}) & 6641 & 293.7 & 44.22 & --  & -- & -- \\
					\midrule
					Positive & \emph{Chunqiu} annals        & 1532 & 19.2  & 12.52  & 10  & 2  & 71   \\
					Positive & \emph{Gongyang} zhuan       & 1776 & 44.5  & 25.05  & 10  & 2  & 671  \\
					Positive & \emph{Zuo} zhuan            & 1547 & 189.1 & 122.20 & 48  & 3  & 2658 \\
					Positive & \emph{Guliang} zhuan        & 1786 & 41.0  & 22.93  & 11  & 2  & 467  \\
					\midrule
					Negative & All exegetical layers       & 9227 & 1014.9 & 109.99 & --  & -- & -- \\
					\midrule
					Negative & Lü Zuqian                   & 241  & 99.0  & 410.70 & 365 & 14 & 1715 \\
					Negative & Kong Yingda                 & 3653 & 483.5 & 132.36 & 83  & 0  & 2243 \\
					Negative & Du Yu                       & 3652 & 218.1 & 59.72  & 49  & 1  & 714  \\
					Negative & Gu Donggao                  & 481  & 39.3  & 81.80  & 62  & 1  & 597  \\
					Negative & Wei Liaoweng                & 1200 & 174.9 & 145.79 & 106 & 10 & 2877 \\
					\bottomrule
				\end{tabular}%
			}
			\caption{Raw record-level character-length statistics before sentence splitting. Character counts are reported in thousands (K). Positive records come from the \emph{Chunqiu} annals and the three traditional \emph{zhuan}, while negative records come from later exegetical layers.}
			\label{tab:raw_length_stats}
		\end{table*}
		
The raw source pools differ substantially in length and discourse style, especially between canonical records and later exegetical materials, which motivates sentence-level segmentation before retrieval construction.		
		
		\begin{figure*}[t]
			\includegraphics[width=\linewidth]{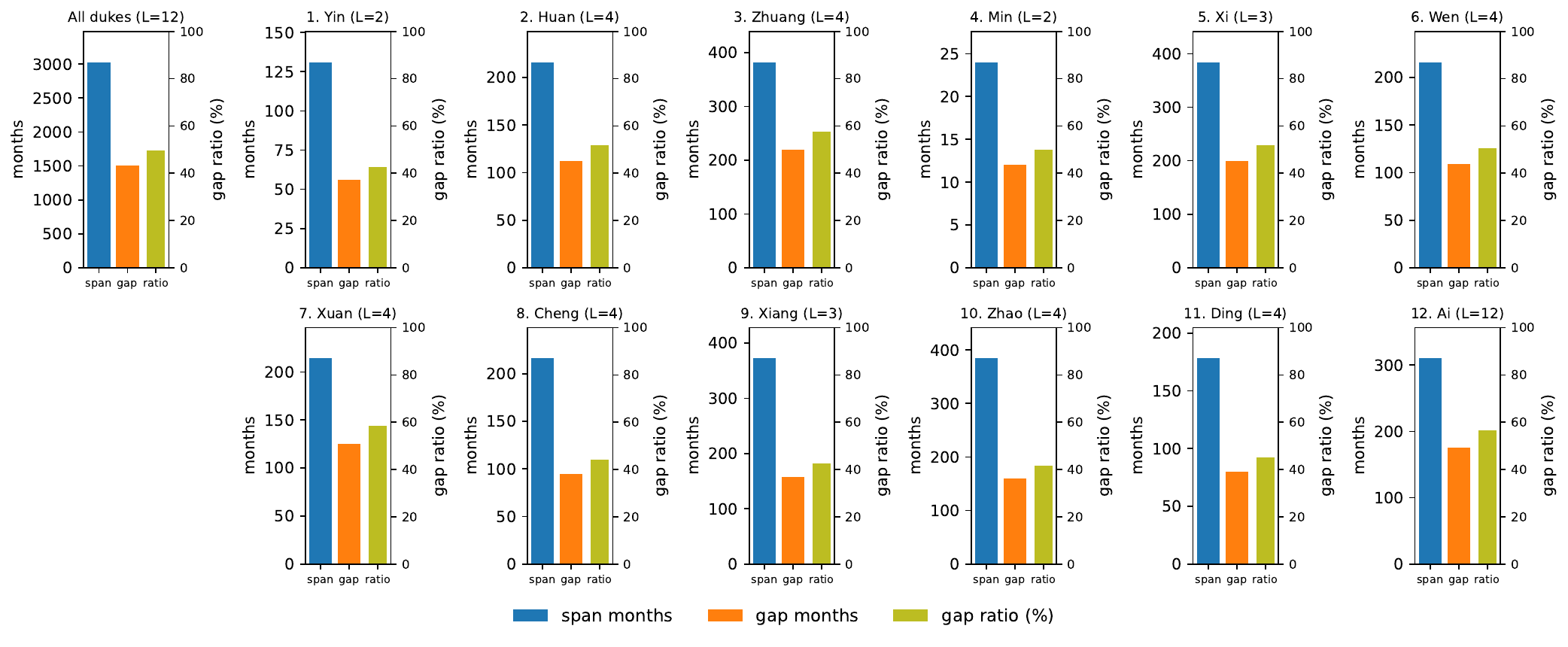}
			\caption{Month-level coverage, gap counts, and gap ratios for the normalized \emph{Chunqiu} timeline (overall and per Lu ruler).}
			\label{fig:appendix_statistic_month}
		\end{figure*}
		
Fig.~\ref{fig:appendix_statistic_month} summarizes month-level coverage and gap months over the normalized \textit{Chunqiu} timeline. The benchmark spans 3036 reign-based months from Duke Yin to Duke Ai, with a substantial proportion of months containing no recorded event, confirming that gap months are a pervasive property of the corpus rather than an edge case.

\subsubsection{Record-level segmentation.}

To construct retrieval units, we convert heterogeneous raw source materials into sentence-level records.
We first use an LLM to propose punctuation and sentence boundaries (\emph{句读}) for each raw passage, and then apply a light rule-based splitter over classical discourse markers such as ``曰'', ``云'', and ``传曰''.
During this step, we also perform minimal normalization to reduce stylistic boilerplate, for example by stripping framing markers such as ``正义曰'' or formulaic quotation headers that do not contribute substantive event content, and by discarding extremely short fragments that contain only a few characters.

Each resulting record inherits the reign-based time key and source metadata of its parent unit, and is assigned a coarse type label: \texttt{event}, \texttt{no\_event}, or \texttt{neg\_comment}.
For months where the annals and the three \emph{zhuan} jointly indicate that nothing was recorded, we additionally synthesize a standardized \texttt{no\_event} record for that time key (e.g., ``鲁隐公元年二月：《春秋》经文及三传于此月无事可书。''),
so that retrieving an empty-month case still requires matching the correct reign and month, rather than collapsing all such queries to a single global ``nothing happened'' entry.

This process yields 20{,}172 records in total, which constitute the retrieval gallery used in our time-aware experiments (Table~\ref{tab:final_split_stats}).

		\subsubsection{Final benchmark splits.}
		
		\begin{table*}[t]
			\centering
			\small
			\resizebox{\textwidth}{!}{%
				\begin{tabular}{lrrrrrrr} 
					\toprule
					Split & \# months & \# records & \# queries & Avg. ground-truth recs/query & \# event recs & \# no-event recs & \# neg.\ comments \\
					\midrule
					Train      & 2424 & 16027 & 13053 & 7.3 & 5360 & 1209 &  9458 \\
					Validation &  295 &  2049 &  1520 & 6.8 &  626 &  152 &  1271 \\
					Test       &  317 &  2096 &  1653 & 7.2 &  782 &  149 &  1165 \\
					\midrule
					Total      & 3036 & 20172 & 16226 & 7.2 & 6768 & 1510 & 11894 \\
					\bottomrule
				\end{tabular}%
			}
			\caption{Final benchmark statistics and splits over month-level time keys, record-level retrieval units, and queries. The ``Avg.\ ground-truth recs/query'' column reports the average number of labeled relevant records per query in each split, and the last three columns break down records by type (\texttt{event}, \texttt{no\_event}, and \texttt{neg\_comment}).}
			\label{tab:final_split_stats}
		\end{table*}

		The benchmark is split at the month level using an approximate 80/10/10 partition, and all records and queries inherit the split of their associated time key to avoid temporal leakage. As shown in Table~\ref{tab:final_split_stats}, the final benchmark contains 3036 month keys, 20{,}172 record-level retrieval units, and 16{,}226 queries, with explicit breakdowns by split and record type.

		\subsubsection{Data sources and licensing.}

		\paragraph{Source \& license.}
		All digitized texts used in ChunQiuTR are retrieved from Chinese Wikisource (\textit{Siku Quanshu} editions). Individual work pages are tagged as public domain (e.g., \texttt{PD-old}), while platform content is provided under CC BY-SA 4.0 and the Wikimedia Terms of Use. To facilitate compliant reuse, we record page revision IDs (\texttt{oldid}) and release the benchmark as derived metadata together with scripts for re-downloading the raw texts.
		
		\begin{table*}[t]
			\centering
			\small
			\setlength{\tabcolsep}{5pt}
			\renewcommand{\arraystretch}{1.12}
            \resizebox{\textwidth}{!}{%
			\begin{tabular}{@{}L{0.26\textwidth} L{0.18\textwidth} L{0.23\textwidth} L{0.33\textwidth}@{}}
				\toprule
				\textbf{Work / layer} &
				\textbf{Role in ChunQiuTR} &
				\textbf{Digital source (edition)} &
				\textbf{License note / release plan} \\
				\midrule
				
				\textit{Chunqiu} (春秋) &
				base corpus &
				\multirow[t]{7}{0.23\textwidth}{Chinese Wikisource (\textit{Siku Quanshu} edition; page revision \texttt{oldid} recorded)} &
				\multirow[t]{7}{0.33\textwidth}{\textbf{Work pages:} tagged \texttt{PD-old}. \textbf{Platform text:} CC~BY-SA~4.0 (Wikimedia Terms of Use).
					We release derived metadata (time keys, alignments, queries/qrels, indices) and scripts to re-fetch raw texts from recorded \texttt{oldid} revisions.} \\
				
				\textit{Zuo zhuan} (左氏傳) &
				base corpus & & \\
				
				\textit{Gongyang zhuan} (公羊傳) &
				base corpus & & \\
				
				\textit{Guliang zhuan} (穀梁傳) &
				base corpus & & \\
				
				Gu Donggao, \textit{Chunqiu Dashibiao} (顧栋高《春秋大事表》) &
				chrono-near paraphrastic negatives & & \\
				
				Wei Liaoweng, \textit{Chunqiu Zuozhuan Yaoyi} (魏了翁《春秋左傳要義》) &
				chrono-near discursive negatives & & \\
				
				\textit{Zuozhuan} annotations / sub-commentaries (注疏; e.g., 音義/杜注/正義/說) &
				lexical-near annotation negatives & & \\
				
				\bottomrule
			\end{tabular}
            }
			\caption{Text sources and licensing. All digitized texts are retrieved from Chinese Wikisource (\textit{Siku Quanshu} editions), with page revision IDs (\texttt{oldid}) recorded for traceability. See Appendix~A for alignment and preprocessing.}
			\label{tab:source_license}
		\end{table*}

		\section{Details of Methods}
        \subsection{Details of Experiment Setting}
        \subsubsection{Model and Training Details}
\label{app:model_details}

We fine-tune two dual-encoder backbones: \textsc{BERT-base-chinese} and \textsc{Qwen3-Embed-0.6B}. 
For \textsc{BERT-base}, we use \texttt{[CLS]} pooling; for \textsc{Qwen3-Embed-0.6B}, we use last-token pooling.

Both backbones are trained with a contrastive retrieval objective using multi-positive supervision and explicit hard-negative training. 
We additionally apply an auxiliary time classification loss over the three discrete factors (gong/year/month), with weight $0.1$ and label smoothing $\epsilon=0.2$. 
For CTD, we enable both the relative temporal bias and the soft absolute temporal context derived from predicted time distributions.

We optimize with AdamW (weight decay $0.01$) and a linear learning-rate schedule with warmup ratio $0.1$. 
For \textsc{BERT-base}, we train for 5 epochs with batch size 64 and learning rate $2\times 10^{-5}$, using maximum query/passage lengths of 64/196. 
For \textsc{Qwen3-Embed-0.6B}, we train for 3 epochs with effective batch size 16 and learning rate $3\times 10^{-6}$, using maximum query/passage lengths of 128/256; we also enable global in-batch negatives.

We select checkpoints by validation \textbf{Recall@1} and report \textbf{Recall@K} and \textbf{MRR@10} on the test split under the same evaluation protocol. During evaluation, both commentary negatives and explicit \texttt{no\_event} records are included in the candidate gallery.

\subsubsection{Training Cost and Computational Resources}
\label{app:compute}

We report the computing infrastructure and approximate GPU-hours for representative fine-tuning runs.

\paragraph{Hardware.}
\textsc{BERT-base-chinese} is trained on a single GPU (either 1$\times$ NVIDIA RTX A6000 or 2$\times$ RTX 3090, depending on availability). 
\textsc{Qwen3-Embed-0.6B} uses multi-GPU distributed training (either 2$\times$ RTX A6000 or 4$\times$ RTX 3090) to support global in-batch negatives.

\paragraph{Training cost.}
Table~\ref{tab:compute_cost} reports representative wall-clock time per run and the corresponding GPU-hours. 
These numbers cover end-to-end fine-tuning with periodic validation; sparse baselines and non-parametric time priors incur negligible training cost.

\begin{table}[t]
\centering
\small
\setlength{\tabcolsep}{6pt}
\resizebox{\linewidth}{!}{%
\begin{tabular}{l l c c c}
\toprule
Backbone & Variant & \#GPU & Time / run & GPU-hours \\
\midrule
\textsc{BERT-base-chinese} & FT baseline & 1 & $\approx$15 min & $\approx$0.25 \\
\textsc{BERT-base-chinese} & CTD (full)  & 1 & $\approx$19 min & $\approx$0.32 \\
\textsc{Qwen3-Embed-0.6B} & FT baseline & 2 & $\approx$45 min & $\approx$1.50 \\
\textsc{Qwen3-Embed-0.6B} & CTD (full)  & 2 & $\approx$45 min & $\approx$1.50 \\
\bottomrule
\end{tabular}
}
\caption{Compute cost for representative fine-tuning runs on ChunQiuTR. Time is wall-clock per run; GPU-hours are computed as (\#GPU)$\times$(time in hours).}
\label{tab:compute_cost}
\end{table}

\paragraph{Hyperparameters.}
We do not perform large-scale hyperparameter sweeps. Instead, we adopt standard fine-tuning settings for each backbone and select the best checkpoint by validation \textbf{Recall@1}.
        
		\subsection{Details of Analysis}
		\subsubsection{Details of Ablation Study}

We further replicate the ablation with \texttt{bert-base-chinese} to examine backbone sensitivity. As shown in Table~\ref{tab:app_ablation_bert}, adding multi-positive retrieval supervision already improves over the FT baseline, while temporal modeling yields substantially larger gains. Among the two temporal modules, the soft absolute temporal context $\mathbf{c}_x$ contributes a stronger overall boost than the relative-time bias $b^{\text{time}}_{ij}$, and combining both gives the best performance. The larger gains on BERT than on stronger embedding backbones suggest that temporal supervision and structured negatives are especially helpful when the base retriever has more room to reduce chrono-near confusions.

		\begin{table}[t]
			\centering
			\small
			\setlength{\tabcolsep}{3pt}
			\renewcommand{\arraystretch}{1.12}
            \resizebox{\linewidth}{!}{%
			\begin{tabular}{l ccc cc}
				\toprule
				\textbf{Variant} &
				$\mathcal{L}_{\text{multi}}$ & $b^{\text{time}}_{ij}$ & $\mathbf{c}_x$ &
				\textbf{R@1} & \textbf{MRR@10} \\
				\midrule
				FT baseline
				& -- & -- & -- &
				\shortstack{0.5088  {\scriptsize ----}} &
				\shortstack{0.5597  {\scriptsize ----}} \\
				
				+ $\mathcal{L}_{\text{multi}}$
				& $\checkmark$ & -- & -- &
				\shortstack{0.5178  {\scriptsize \up{+0.0090}}} &
				\shortstack{0.5685  {\scriptsize \up{+0.0088}}} \\
				
				+ Bias
				& $\checkmark$ & $\checkmark$ & -- &
				\shortstack{0.5384  {\scriptsize \up{+0.0296}}} &
				\shortstack{0.5776  {\scriptsize \up{+0.0179}}} \\
				
				+ Ctx
				& $\checkmark$ & -- & $\checkmark$ &
				\shortstack{0.5620  {\scriptsize \up{+0.0532}}} &
				\shortstack{0.5961  {\scriptsize \up{+0.0364}}} \\
				
				\textbf{Full (Ours)}
				& $\checkmark$ & $\checkmark$ & $\checkmark$ &
				\shortstack{0.5826  {\scriptsize \up{+0.0738}}} &
				\shortstack{0.6193  {\scriptsize \up{+0.0596}}} \\
				\bottomrule
			\end{tabular}%
            }
			\caption{
				Ablation on the test set with \texttt{bert-base-chinese}.
			}
			\label{tab:app_ablation_bert}
		\end{table}
		
		\begin{table*}[t]
			\centering
			\small
			\setlength{\tabcolsep}{3pt}
			\renewcommand{\arraystretch}{1.12}
			\resizebox{\textwidth}{!}{%
				\begin{tabular}{l cc cc cc cc cc cc cc }
					\toprule
					\multirow{2}{*}{\textbf{Method}} &
					\multicolumn{6}{c}{\textbf{Including pure no-event queries} (fix \texttt{neg}=0, \texttt{ne}=1)} &
					\multicolumn{6}{c}{\textbf{Hard-negative robustness} (fix \texttt{ne=1, dq=1})} \\
					\cmidrule(lr){2-7}\cmidrule(lr){8-13}
					& \multicolumn{2}{c}{\textbf{\texttt{dq=1}}} & \multicolumn{2}{c}{\textbf{\texttt{dq=0}}} & \multicolumn{2}{c}{\textbf{$\Delta$}} 
					& \multicolumn{2}{c}{\textbf{\texttt{neg=0}}} & \multicolumn{2}{c}{\textbf{\texttt{neg=1}}} & \multicolumn{2}{c}{\textbf{$\Delta$}} \\
					\cmidrule(lr){2-3}\cmidrule(lr){4-5}\cmidrule(lr){6-7}
					\cmidrule(lr){8-9}\cmidrule(lr){10-11}\cmidrule(lr){12-13}
					& R@1 & MRR@10 & R@1 & MRR@10 & R@1 & MRR@10
					& R@1 & MRR@10 & R@1 & MRR@10 & R@1 & MRR@10 \\
					\midrule
					BM25 & 0.240 & 0.296 & 0.418 & 0.470 & \up{+0.178} & \up{+0.173} & 0.240 & 0.296 & 0.228 & 0.282 & \down{-0.012} & \down{-0.014} \\
					ColBERT-LFM2$_{(\text{ZS})}$ & 0.220 & 0.277 & 0.333 & 0.386 & \up{+0.113} & \up{+0.109} & 0.220 & 0.277 & 0.222 & 0.278 & \up{+0.002} & \up{+0.001} \\
					BERT-base$_{(\text{FT})}$ & 0.378 & 0.439 & 0.560 & 0.606 & \up{+0.182} & \up{+0.167} & 0.378 & 0.439 & 0.306 & 0.374 & \down{-0.072} & \down{-0.065} \\
					\textbf{\model$_{\text{BERT-base}}$ (Ours)} & 0.407 & 0.456 & 0.583 & 0.619 & \up{+0.175} & \up{+0.163} & 0.407 & 0.456 & 0.407 & 0.456 & {\scriptsize 0.000} & {\scriptsize 0.000} \\
					\midrule
					E5-mistral-7B$_{(\text{ZS})}$ & 0.062 & 0.093 & 0.225 & 0.270 & \up{+0.163} & \up{+0.177} & 0.062 & 0.093 & 0.054 & 0.082 & \down{-0.008} & \down{-0.011} \\
					Qwen3-Embed-0.6B$_{(\text{FT})}$ & 0.396 & 0.434 & 0.577 & 0.605 & \up{+0.181} & \up{+0.171} & 0.396 & 0.434 & 0.396 & 0.433 & {\scriptsize 0.000} & \down{-0.001} \\
					\textbf{\model$_{\text{Qwen3-Embed-0.6B}}$ (Ours)} & 0.420 & 0.457 & 0.594 & 0.621 & \up{+0.174} & \up{+0.164} & 0.420 & 0.457 & 0.418 & 0.455 & \down{-0.002} & \down{-0.002} \\
					\bottomrule
				\end{tabular}%
			}
			\caption{
No-event and hard-negative behavior under protocol variations on the test set.
Left: \texttt{dq=1} vs.\ \texttt{dq=0} (fix \texttt{ne}=1, \texttt{neg}=0), where \texttt{dq} drops or keeps pure \texttt{no\_event} queries.
Right: \texttt{neg=0} vs.\ \texttt{neg=1} (fix \texttt{ne}=1, \texttt{dq}=1), where \texttt{neg} injects \texttt{neg\_comment} passages into the gallery.
$\Delta$ denotes the within-method change.
}
			\label{tab:noevent_hardneg_behavior}
		\end{table*}
		
		\subsubsection{No-event and Hard-negative Behavior}

Table~\ref{tab:noevent_hardneg_behavior} probes two protocol switches: whether to keep pure \texttt{no\_event} queries (\texttt{dq}) and whether to inject \texttt{neg\_comment} passages as chrono-near hard negatives (\texttt{neg}). Including pure no-event queries consistently raises scores across all methods, indicating that empty-month retrieval is substantially easier and should be controlled by protocol. By contrast, injecting \texttt{neg\_comment} passages exposes genuine robustness differences: the BERT FT baseline drops noticeably, whereas \model$_{\text{BERT-base}}$ remains stable, and both Qwen-based systems change only marginally. Overall, the benchmark contains both an easier no-event regime and a harder exegetical hard-negative regime, with CTD improving robustness especially for weaker backbones.

        
        \subsubsection{Full Protocol Grid Results}
\label{sec:app_full_grid}

We report results under all valid combinations of the three evaluation switches: \texttt{neg} (whether \texttt{neg\_comment} passages are included in the gallery), \texttt{ne} (whether explicit \texttt{no\_event} records are included), and \texttt{dq} (whether pure \texttt{no\_event} queries are dropped). Modes are denoted as \texttt{neg\{0/1\}\_ne\{0/1\}\_dq\{0/1\}}. When \texttt{ne=0}, pure \texttt{no\_event} queries are ill-defined for retrieval, so only \texttt{dq=1} is reported.

Tables~\ref{tab:app_full_grid_ours} and~\ref{tab:app_full_grid_bm25} summarize validation and test results for all queries, and separately for the point and window families. Two trends are consistent across settings: keeping pure \texttt{no\_event} queries (\texttt{dq=0}) raises aggregate scores, whereas injecting \texttt{neg\_comment} passages (\texttt{neg=1}) yields a harder and more realistic gallery. The full grid is therefore intended mainly as a protocol reference and robustness diagnostic.
        
        \paragraph{Full-grid results.}
        Tables~\ref{tab:app_full_grid_ours} and~\ref{tab:app_full_grid_bm25} report Recall@K, MRR@10, and nDCG@10 on validation and test under each mode.
        We report results on \texttt{all} queries, and also stratify by \texttt{point} and \texttt{window} families (corresponding to our point-/gap-like vs.\ window-style temporal probes in the main paper).
        
        \begin{table*}[t]
            \centering
            \scriptsize
            \setlength{\tabcolsep}{3pt}
            \renewcommand{\arraystretch}{1.08}
            \resizebox{\textwidth}{!}{%
            \begin{tabular}{llccccc ccccc}
                \toprule
                \multirow{2}{*}{\textbf{Mode}} & \multirow{2}{*}{\textbf{Family}} &
                \multicolumn{5}{c}{\textbf{Validation}} &
                \multicolumn{5}{c}{\textbf{Test}} \\
                \cmidrule(lr){3-7}\cmidrule(lr){8-12}
                & & R@1 & R@5 & R@10 & MRR@10 & nDCG@10
                  & R@1 & R@5 & R@10 & MRR@10 & nDCG@10 \\
                \midrule
                \texttt{neg0\_ne0\_dq1} & all    & 0.0407 & 0.1200 & 0.2232 & 0.0798 & 0.0635 & 0.0654 & 0.1466 & 0.2042 & 0.1014 & 0.0677 \\
                \texttt{neg0\_ne0\_dq1} & point  & 0.0375 & 0.1148 & 0.2459 & 0.0775 & 0.0756 & 0.0710 & 0.1696 & 0.2387 & 0.1156 & 0.0934 \\
                \texttt{neg0\_ne0\_dq1} & window & 0.0430 & 0.1239 & 0.2065 & 0.0815 & 0.0546 & 0.0610 & 0.1283 & 0.1768 & 0.0902 & 0.0472 \\
                \midrule
                \texttt{neg0\_ne1\_dq0} & all    & 0.6329 & 0.6697 & 0.7191 & 0.6522 & 0.4921 & 0.5935 & 0.6497 & 0.6945 & 0.6206 & 0.4588 \\
                \texttt{neg0\_ne1\_dq0} & point  & 0.5200 & 0.5440 & 0.6057 & 0.5347 & 0.5380 & 0.4922 & 0.5307 & 0.5869 & 0.5138 & 0.5103 \\
                \texttt{neg0\_ne1\_dq0} & window & 0.7860 & 0.8403 & 0.8729 & 0.8115 & 0.4300 & 0.7341 & 0.8150 & 0.8439 & 0.7689 & 0.3871 \\
                \midrule
                \texttt{neg0\_ne1\_dq1} & all    & 0.4534 & 0.5050 & 0.5784 & 0.4806 & 0.2385 & 0.4197 & 0.4956 & 0.5602 & 0.4565 & 0.2223 \\
                \texttt{neg0\_ne1\_dq1} & point  & 0.0164 & 0.0656 & 0.1920 & 0.0466 & 0.0532 & 0.0375 & 0.1105 & 0.2170 & 0.0784 & 0.0719 \\
                \texttt{neg0\_ne1\_dq1} & window & 0.7745 & 0.8279 & 0.8623 & 0.7996 & 0.3747 & 0.7230 & 0.8013 & 0.8326 & 0.7565 & 0.3417 \\
                \midrule
                \texttt{neg1\_ne0\_dq1} & all    & 0.0317 & 0.1101 & 0.2173 & 0.0709 & 0.0600 & 0.0593 & 0.1344 & 0.1998 & 0.0951 & 0.0647 \\
                \texttt{neg1\_ne0\_dq1} & point  & 0.0328 & 0.1077 & 0.2436 & 0.0725 & 0.0729 & 0.0690 & 0.1637 & 0.2367 & 0.1127 & 0.0912 \\
                \texttt{neg1\_ne0\_dq1} & window & 0.0310 & 0.1119 & 0.1979 & 0.0697 & 0.0505 & 0.0516 & 0.1111 & 0.1706 & 0.0812 & 0.0437 \\
                \midrule
                \texttt{neg1\_ne1\_dq0} & all    & 0.6329 & 0.6691 & 0.7184 & 0.6520 & 0.4914 & 0.5923 & 0.6485 & 0.6927 & 0.6194 & 0.4575 \\
                \texttt{neg1\_ne1\_dq0} & point  & 0.5200 & 0.5429 & 0.6046 & 0.5344 & 0.5373 & 0.4912 & 0.5307 & 0.5858 & 0.5129 & 0.5094 \\
                \texttt{neg1\_ne1\_dq0} & window & 0.7860 & 0.8403 & 0.8729 & 0.8115 & 0.4291 & 0.7327 & 0.8121 & 0.8410 & 0.7674 & 0.3854 \\
                \midrule
                \texttt{neg1\_ne1\_dq1} & all    & 0.4534 & 0.5040 & 0.5774 & 0.4803 & 0.2374 & 0.4180 & 0.4939 & 0.5576 & 0.4548 & 0.2205 \\
                \texttt{neg1\_ne1\_dq1} & point  & 0.0164 & 0.0632 & 0.1897 & 0.0459 & 0.0519 & 0.0355 & 0.1105 & 0.2150 & 0.0767 & 0.0701 \\
                \texttt{neg1\_ne1\_dq1} & window & 0.7745 & 0.8279 & 0.8623 & 0.7996 & 0.3737 & 0.7214 & 0.7981 & 0.8294 & 0.7549 & 0.3399 \\
                \bottomrule
            \end{tabular}%
            }
            \caption{Full protocol grid results for \textbf{CTD}-\textsc{Qwen3-Embed-0.6B}. Mode names follow \texttt{neg}/\texttt{ne}/\texttt{dq} as defined in Section~\ref{sec:app_full_grid}.}
            \label{tab:app_full_grid_ours}
        \end{table*}
        
        \begin{table*}[t]
            \centering
            \scriptsize
            \setlength{\tabcolsep}{3pt}
            \renewcommand{\arraystretch}{1.08}
            \resizebox{\textwidth}{!}{%
            \begin{tabular}{llccccc ccccc}
                \toprule
                \multirow{2}{*}{\textbf{Mode}} & \multirow{2}{*}{\textbf{Family}} &
                \multicolumn{5}{c}{\textbf{Validation}} &
                \multicolumn{5}{c}{\textbf{Test}} \\
                \cmidrule(lr){3-7}\cmidrule(lr){8-12}
                & & R@1 & R@5 & R@10 & MRR@10 & nDCG@10
                  & R@1 & R@5 & R@10 & MRR@10 & nDCG@10 \\
                \midrule
                \texttt{neg0\_ne0\_dq1} & all    & 0.0159 & 0.0456 & 0.0952 & 0.0331 & 0.0205 & 0.0096 & 0.0366 & 0.0672 & 0.0234 & 0.0192 \\
                \texttt{neg0\_ne0\_dq1} & point  & 0.0258 & 0.0468 & 0.0937 & 0.0382 & 0.0265 & 0.0079 & 0.0394 & 0.0690 & 0.0226 & 0.0235 \\
                \texttt{neg0\_ne0\_dq1} & window & 0.0086 & 0.0448 & 0.0964 & 0.0294 & 0.0161 & 0.0110 & 0.0344 & 0.0657 & 0.0240 & 0.0158 \\
                \midrule
                \texttt{neg0\_ne1\_dq0} & all    & 0.4474 & 0.5678 & 0.5980 & 0.4989 & 0.3794 & 0.4180 & 0.5408 & 0.5735 & 0.4696 & 0.3543 \\
                \texttt{neg0\_ne1\_dq0} & point  & 0.4434 & 0.4709 & 0.4789 & 0.4556 & 0.4612 & 0.4152 & 0.4475 & 0.4506 & 0.4284 & 0.4339 \\
                \texttt{neg0\_ne1\_dq0} & window & 0.4527 & 0.6992 & 0.7597 & 0.5576 & 0.2683 & 0.4220 & 0.6705 & 0.7442 & 0.5269 & 0.2436 \\
                \midrule
                \texttt{neg0\_ne1\_dq1} & all    & 0.2629 & 0.4097 & 0.4464 & 0.3261 & 0.1398 & 0.2400 & 0.3709 & 0.4127 & 0.2963 & 0.1218 \\
                \texttt{neg0\_ne1\_dq1} & point  & 0.0000 & 0.0000 & 0.0000 & 0.0000 & 0.0000 & 0.0000 & 0.0000 & 0.0000 & 0.0000 & 0.0000 \\
                \texttt{neg0\_ne1\_dq1} & window & 0.4561 & 0.7108 & 0.7745 & 0.5658 & 0.2426 & 0.4304 & 0.6651 & 0.7402 & 0.5314 & 0.2184 \\
                \midrule
                \texttt{neg1\_ne0\_dq1} & all    & 0.0000 & 0.0079 & 0.0169 & 0.0043 & 0.0043 & 0.0000 & 0.0052 & 0.0122 & 0.0023 & 0.0023 \\
                \texttt{neg1\_ne0\_dq1} & point  & 0.0000 & 0.0117 & 0.0258 & 0.0073 & 0.0076 & 0.0000 & 0.0059 & 0.0178 & 0.0033 & 0.0037 \\
                \texttt{neg1\_ne0\_dq1} & window & 0.0000 & 0.0052 & 0.0103 & 0.0021 & 0.0019 & 0.0000 & 0.0047 & 0.0078 & 0.0015 & 0.0012 \\
                \midrule
                \texttt{neg1\_ne1\_dq0} & all    & 0.4303 & 0.5480 & 0.5908 & 0.4809 & 0.3654 & 0.3962 & 0.5209 & 0.5620 & 0.4487 & 0.3404 \\
                \texttt{neg1\_ne1\_dq0} & point  & 0.4286 & 0.4571 & 0.4754 & 0.4415 & 0.4494 & 0.3965 & 0.4350 & 0.4495 & 0.4125 & 0.4214 \\
                \texttt{neg1\_ne1\_dq0} & window & 0.4326 & 0.6713 & 0.7473 & 0.5343 & 0.2515 & 0.3960 & 0.6402 & 0.7182 & 0.4989 & 0.2280 \\
                \midrule
                \texttt{neg1\_ne1\_dq1} & all    & 0.2530 & 0.3948 & 0.4395 & 0.3137 & 0.1314 & 0.2277 & 0.3560 & 0.3988 & 0.2823 & 0.1151 \\
                \texttt{neg1\_ne1\_dq1} & point  & 0.0000 & 0.0000 & 0.0000 & 0.0000 & 0.0000 & 0.0000 & 0.0000 & 0.0000 & 0.0000 & 0.0000 \\
                \texttt{neg1\_ne1\_dq1} & window & 0.4389 & 0.6850 & 0.7625 & 0.5442 & 0.2279 & 0.4085 & 0.6385 & 0.7152 & 0.5063 & 0.2064 \\
                \bottomrule
            \end{tabular}%
            }
            \caption{Full protocol grid results for BM25, reported in the same format as Table~\ref{tab:app_full_grid_ours}.}
            \label{tab:app_full_grid_bm25}
        \end{table*}
        
        \paragraph{What this grid clarifies.}
        Two takeaways are particularly relevant for interpreting aggregate scores.
        First, \texttt{dq=0} (keeping pure no-event queries) can noticeably inflate overall metrics compared to \texttt{dq=1}, motivating our practice of reporting both settings: \texttt{dq=0} reflects the benchmark’s intended scope (event months \emph{and} explicit empty months), while \texttt{dq=1} isolates event-seeking behavior.
        Second, injecting exegetical hard negatives (\texttt{neg=1}) is a strictly harder and more realistic gallery setting; models that remain stable between \texttt{neg=0} and \texttt{neg=1} exhibit stronger robustness to chrono-near confounds from commentarial material.

        \subsubsection{Top-1 Near-miss Failure Cases}
\label{sec:app_failure_cases}

\begin{figure*}[t]
            \centering
            \includegraphics[width=\textwidth]{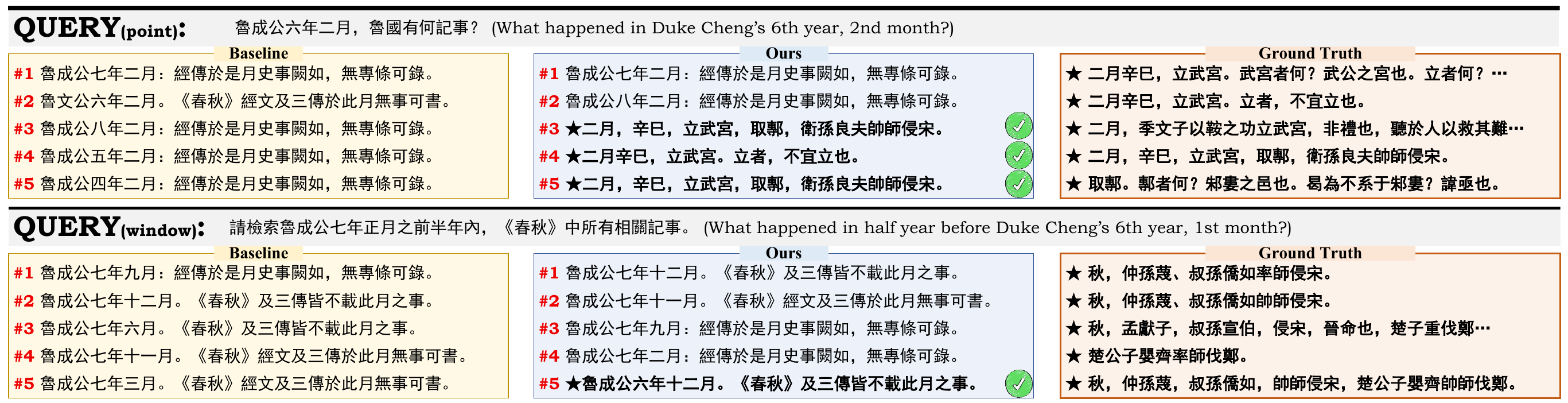}
            \caption{
            Top-1 near-miss cases on ChunQiuTR (test set).
            We show the top--5 results from a baseline (left) and ours (middle), with the ground-truth set (right).
            $\star$ marks ground-truth passages; \checkmark indicates a ground-truth hit in top--5.
            }
            \label{fig:app_top1_failures}
        \end{figure*}

Figure~\ref{fig:app_top1_failures} shows two representative near-miss cases on the test set: our retriever fails to place the correct passage at rank~1, but still retrieves at least one ground-truth passage within the top--5, whereas the baseline fails to surface any ground-truth evidence. In both cases, a key confounder is the highly reusable \texttt{no\_event}-style wording and its chrono-near reoccurrence across adjacent months, which can trigger top-rank swaps. These examples suggest that the remaining errors are often ordering errors under strong lexical or temporal confounders, rather than complete retrieval failure.
        
        \subsubsection{Qualitative demo: reasoning traces vs.\ evidence grounding}
\label{sec:app_rag_demo}

\begin{figure}[t]
			\includegraphics[width=0.85\columnwidth]{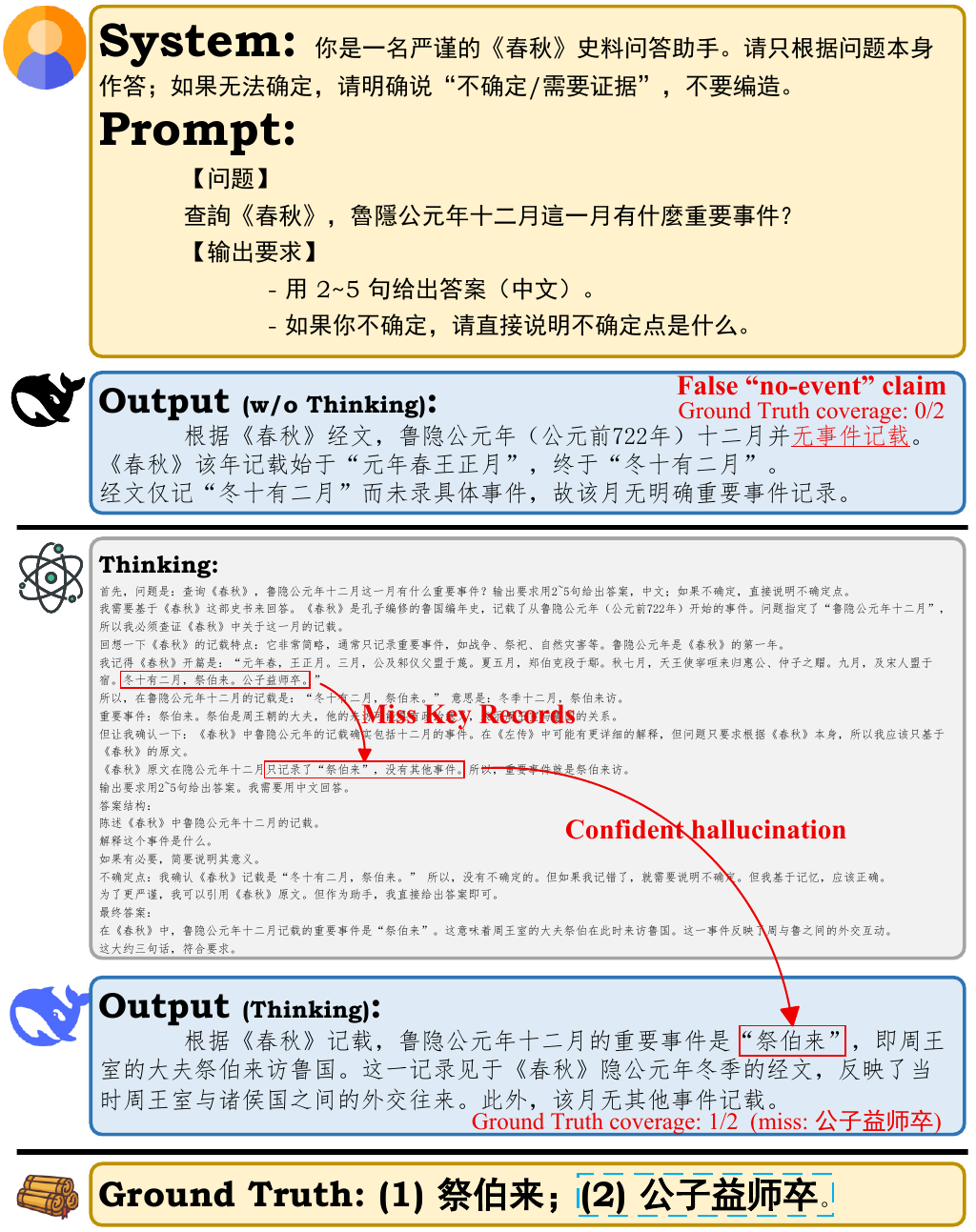}
			\caption{Online LLM without evidence grounding on a month-level point query from ChunQiuTR. For the query ``鲁隐公元年十二月'', the gold month contains two entries (\emph{祭伯来} and \emph{公子益师卒}). Without evidence, the model either predicts an empty month or returns an incomplete answer.}
            \label{fig:demo_norag_deepseek}
		\end{figure}

\begin{figure}[t]
			\includegraphics[width=0.9\columnwidth]{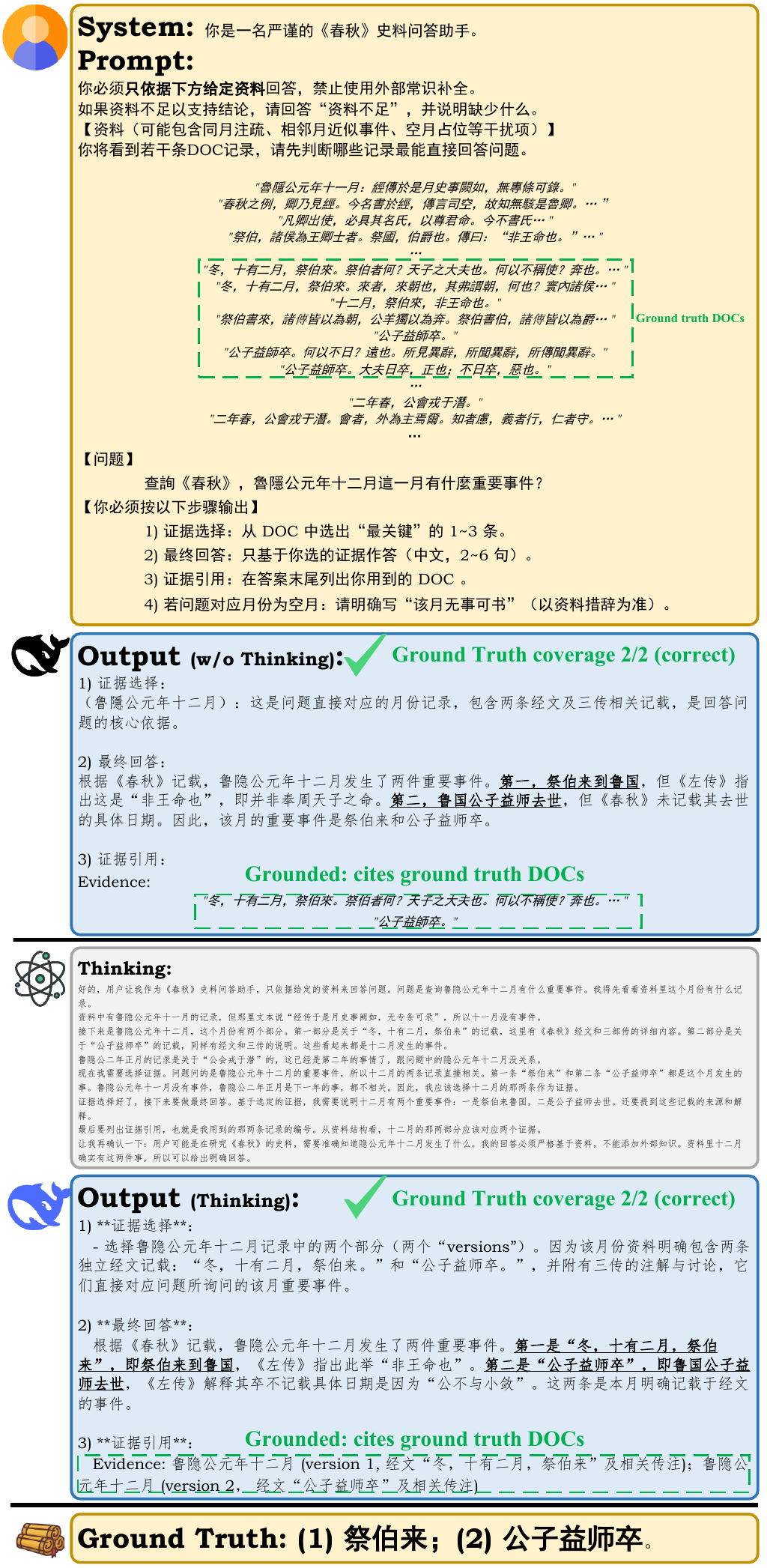}
			\caption{Evidence-bounded RAG for the same query as Fig.~\ref{fig:demo_norag_deepseek}. With a small evidence pack containing the gold month records and confusable materials, the model recovers both gold entries and grounds the answer in cited evidence.}
            \label{fig:demo_rag_ours}

		\end{figure}

Figures~\ref{fig:demo_norag_deepseek} and~\ref{fig:demo_rag_ours} compare an online LLM on the same month-level point query with and without evidence grounding. Without retrieved evidence, the model either predicts an empty month or produces an incomplete answer even when a reasoning trace is enabled. When given an evidence pack that contains the gold month records together with confusable materials, the same model recovers both gold entries and grounds the answer in cited evidence. These examples suggest that longer reasoning traces alone do not ensure month-level completeness, whereas explicit evidence binding substantially improves temporal faithfulness.

    		\subsection{Details of Compared Methods}
\label{sec:app_details_comp_baseline}

\subsubsection{Sparse retrieval.}
We compare against a sparse family including a classical lexical retriever, a simple temporal re-ranking variant, and two inference-free neural sparse retrievers, all evaluated under the same sparse retrieval protocol.
\begin{itemize}
    \item \textbf{BM25}~\citep{Robertson2009BM25}: standard lexical term-matching baseline.
    \item \textbf{BM25+TimeKDE}: BM25 with a non-parametric temporal re-ranking prior over regnal-month indices, following classical TIR-style temporal priors~\citep{li2003timebasedlm}.
    \item \textbf{SPLADE-IDF}$_{(\text{ZS})}$~\citep{geng2025competitivesearchrelevanceinferencefree}: inference-free neural sparse retriever used zero-shot.
    \item \textbf{SPLADE-$\ell_0$}$_{(\text{ZS})}$~\citep{Shen2025sigirExploring}: sparsity-controlled neural sparse retriever used zero-shot.
\end{itemize}

\subsubsection{Fusion / late-interaction retrieval.}
We further compare against two multi-vector late-interaction retrievers, both used in a zero-shot setting.
\begin{itemize}
    \item \textbf{ColBERT-JINA}$_{(\text{ZS})}$~\citep{xiao-etal-2024-jina}: ColBERT-style token-interaction retriever.
    \item \textbf{ColBERT-LFM2}$_{(\text{ZS})}$~\citep{amini2025lfm2technicalreport}: late-interaction retriever with longer-context and multi-scale representations.
\end{itemize}

\subsubsection{Dense retrieval, encoder-based.}
For encoder-based dense retrieval, we compare against single-vector dual-encoder models used either zero-shot or fine-tuned on ChunQiuTR.
\begin{itemize}
    \item \textbf{GTR-T5-Base / Sentence-T5-Base}$_{(\text{ZS})}$~\citep{ni-etal-2022-large-GTR,ni-etal-2022-sentence-t5}: T5-based dense retrievers used zero-shot.
    \item \textbf{mE5-Large / mE5-Large-ins}$_{(\text{ZS})}$~\citep{wang2024multilinguale5textembeddings}: multilingual E5 retrievers used zero-shot.
    \item \textbf{GTE-Large}$_{(\text{ZS})}$~\citep{li2023generaltextembeddingsmultistage}: general-purpose dense embedding baseline.
    \item \textbf{BGE-Large-v1.5 / BGE-M3}$_{(\text{ZS})}$~\citep{bge_embedding,chen-etal-2024-m3}: strong multilingual dense embedding baselines.
    \item \textbf{BERT-base}$_{(\text{FT})}$~\citep{devlin-etal-2019-bert}: Chinese BERT dual-encoder fine-tuned on ChunQiuTR without explicit time modeling.
\end{itemize}

\subsubsection{Dense retrieval, LM-based embeddings.}
We also compare against LM-based dense embedding models, including both zero-shot and task-adapted variants.
\begin{itemize}
    \item \textbf{GTE-Qwen2-1.5B / E5-Mistral-7B}$_{(\text{ZS})}$~\citep{wang-etal-2024-improving-text}: LLM-scale embedding baselines used zero-shot.
    \item \textbf{PQR (Qwen2.5-7B / Qwen3-8B)}$_{(\text{re})}$~\citep{kang-etal-2025-pqr}: training-free retrieval framework based on LLM-generated pseudo-queries.
    \item \textbf{Qwen3-Embed-0.6B / 4B}$_{(\text{ZS})}$~\citep{zhang2025qwen3embeddingadvancingtext}: dedicated Qwen3 embedding models used zero-shot.
    \item \textbf{Qwen3-Embed-0.6B}$_{(\text{FT})}$~\citep{zhang2025qwen3embeddingadvancingtext}: task-adapted dense dual-encoder baseline without explicit time modeling.
\end{itemize}

\subsubsection{Time-aware auxiliary variants.}
Beyond BM25+TimeKDE, we report two lightweight temporal extensions for single-vector dense retrievers.
\begin{itemize}
    \item \textbf{TempDate}: auxiliary time-key prediction over $(\text{gong},\text{year},\text{month})$ during training, discarded at inference time~\citep{wang2023BiTimeBERT,dhingra-etal-2022-time}.
    \item \textbf{TempDate-Smooth}: TempDate with neighbor-aware smoothing over adjacent ordered time keys~\citep{pmlr-v202-yeche23a}.
\end{itemize}

\subsection{Cross-Corpus Pilot on \textit{Zizhi Tongjian}}
\label{app:zztj-pilot}

To probe whether the temporal-consistency bias learned on ChunQiuTR transfers beyond the \textit{Spring and Autumn Annals}, we conduct a lightweight cross-corpus evaluation on two processed subsets from \textit{Zizhi Tongjian} (\textit{Qi Ji} and \textit{Jin Ji}). As an annalistic general history, \textit{Zizhi Tongjian} also records events under traditional reign-based, non-Gregorian temporal expressions, making it a natural out-of-domain probe for month-keyed retrieval.

This pilot preserves the core month-key retrieval idea of ChunQiuTR but is intentionally lighter than the full benchmark. We retain event-bearing lines as retrieval units, group them by normalized month keys derived from the available reign/year/month fields, and instantiate one point-style query for each unique month key using a traditional reign-year template. No target-corpus training is performed. Unlike the full ChunQiuTR benchmark, this pilot does not reconstruct explicit \texttt{no\_event} placeholders, commentary-derived hard negatives, or the full point/gap/window query families, and should therefore be interpreted as a transfer probe rather than a second benchmark.

\paragraph{Subset statistics.}
Table~\ref{tab:zztj-subsets} summarizes the two processed subsets used in this pilot. \textit{Qi Ji} contains 268 records and 92 month-level queries, while \textit{Jin Ji} contains 820 records and 119 queries. The two slices cover distinct reign periods and remain clearly separate from the ChunQiuTR source corpus.

\begin{table*}[t]
\centering
\small
\setlength{\tabcolsep}{5pt}
\begin{tabular}{lcccc}
\toprule
Subset & Approx. coverage & Representative reign titles & Records & Queries \\
\midrule
Qi Ji (part)  & 479--489 CE & 建元, 永明 & 268 & 92 \\
Jin Ji (part) & 265--279 CE & 泰始, 咸宁 & 820 & 119 \\
\bottomrule
\end{tabular}
\caption{Basic statistics of the processed \textit{Zizhi Tongjian} subsets used in the cross-corpus pilot.}
\label{tab:zztj-subsets}
\end{table*}

\paragraph{Transfer results.}
Table~\ref{tab:zztj-transfer-full} reports retrieval performance on the two subsets. We compare a zero-shot Qwen3-Embed-0.6B encoder, a ChunQiuTR fine-tuned dense baseline, and our CTD-enhanced retriever. Across both subsets, CTD consistently improves MRR and R@1 over the fine-tuned baseline without any target-corpus retraining; on \textit{Qi Ji} it also improves R@5 and R@10, while on \textit{Jin Ji} it matches the fine-tuned baseline on higher-recall metrics.

\begin{table*}[t]
\centering
\small
\setlength{\tabcolsep}{4pt}
\begin{tabular}{lcccc|cccc}
\toprule
\multirow{2}{*}{Model / Setting}
& \multicolumn{4}{c|}{Qi Ji (part)}
& \multicolumn{4}{c}{Jin Ji (part)} \\
& MRR & R@1 & R@5 & R@10 & MRR & R@1 & R@5 & R@10 \\
\midrule
Qwen3-Embed-0.6B (ZS)
& 0.0692 & 0.0217 & 0.0870 & 0.1413
& 0.0691 & 0.0420 & 0.0756 & 0.1345 \\
Qwen3-Embed-0.6B (FT baseline)
& 0.2081 & 0.1848 & 0.2174 & 0.2391
& 0.1598 & 0.1345 & 0.1849 & 0.1849 \\
CTD (Ours)
& \textbf{0.2304} & \textbf{0.2065} & \textbf{0.2391} & \textbf{0.2717}
& \textbf{0.1751} & \textbf{0.1597} & 0.1849 & 0.1849 \\
\bottomrule
\end{tabular}
\caption{Cross-corpus pilot results on two processed \textit{Zizhi Tongjian} subsets. No target-corpus training is performed.}
\label{tab:zztj-transfer-full}
\end{table*}

\paragraph{Discussion.}
Although this transfer setting is lighter than the full ChunQiuTR benchmark, the overall trend is consistent with our main findings: the temporal-consistency bias learned on ChunQiuTR transfers beyond in-domain fitting and continues to help distinguish chrono-near but temporally mismatched evidence. At the same time, the gains are smaller than those observed on the source benchmark, which is expected given both the domain shift and the simplified evaluation protocol. We therefore view this pilot as evidence of promising cross-corpus transfer, rather than as a replacement for a fully reconstructed \textit{Zizhi Tongjian}-specific benchmark.

\subsection{Alignment Audits and Reliability}
\label{sec:appendix_alignment_audit}

To improve the auditability of ChunQiuTR, we summarize here the two LLM-assisted curation stages and the corresponding human-verification statistics. As clarified in the revised main text, ChunQiuTR is \emph{not} an AI-generated dataset: the retrieval gallery is derived from authentic historical sources, the queries are instantiated from a small set of manually written templates, and LLMs are used only to propose candidate splits or candidate alignments during curation. They are never used to generate, rewrite, translate, or paraphrase historical content, and only human-approved results enter the final benchmark.

\paragraph{A. Time-key normalization and manual verification.}
The corpus follows an implicit Lu-state reign calendar, normalized as month-level keys
\[
\tau=(\text{gong},\text{year},\text{month}).
\]
However, many passages do not explicitly contain a complete \((\text{gong},\text{year},\text{month})\) triple. Instead, the ruling duke and/or regnal year must often be recovered from annalistic structure, discourse continuity, and neighboring entries rather than extracted as standalone temporal mentions. For this reason, the mapping from original records to normalized time keys is manually verified during dataset construction, rather than delegated to a fully automatic temporal-expression extractor. Representative examples of reign-key propagation and normalization are provided in Appendix~\ref{sec:reign-time-key-details}.

\paragraph{B. Audit of multi-event splitting.}
A first LLM-assisted step is used when a single month-level segment contains more than one historical event. In these cases, the model is asked only to propose candidate event-level groupings under a fixed time key; all such proposals are then manually reviewed and corrected if necessary.

Table~\ref{tab:audit_multi_event_split} reports the corresponding audit statistics. Among 1,533 non-empty months, 558 contain multiple events (36.41\%). After LLM candidate grouping, only 63 multi-event months required additional human correction, corresponding to 11.29\% of multi-event months and 4.11\% of all non-empty months. The remaining 495 multi-event months were accepted without change (88.71\% direct acceptance among multi-event cases). These statistics suggest that LLM proposal is useful for reducing manual effort, while final segmentation quality remains controlled by explicit human review.

\begin{table}[t]
\centering
\small
\setlength{\tabcolsep}{6pt}
\begin{tabular}{lr}
\toprule
Item & Value \\
\midrule
Total non-empty months & 1,533 \\
Months containing multiple events & 558 \\
Fraction multi-event & 36.41\% \\
Extra human corrections & 63 \\
Correction rate among multi-event months & 11.29\% \\
Corrections among all non-empty months & 4.11\% \\
Accepted without change & 495 \\
Direct acceptance rate & 88.71\% \\
\bottomrule
\end{tabular}
\caption{Audit statistics for multi-event splitting (LLM proposals + human verification).}
\label{tab:audit_multi_event_split}
\end{table}

\paragraph{Common correction patterns.}
Manual corrections in this stage mainly fall into a small number of recurrent categories: (i) \emph{boundary shift}, where the candidate split cuts too early or too late and therefore mixes material from adjacent events; (ii) \emph{inappropriate merge}, where two historically distinct events are grouped together because they share a compact annalistic sentence; and (iii) \emph{inappropriate split}, where commentary fragments that should remain attached to one event are separated into different candidate groups. In all such cases, the final retained record structure is determined by human verification.

\paragraph{C. Audit of later-commentary alignment.}
A second LLM-assisted step is used when aligning later historiographical or commentarial materials to original Chunqiu records. These later sources often refer to canonical events through highly compressed paraphrases, lexical reformulations, or short subtitles rather than direct quotation. We therefore use an LLM only to propose candidate matched passages, after which human verification determines whether the candidate alignment is accepted into the benchmark as a chrono-near confusable negative.

Table~\ref{tab:audit_commentary_alignment} reports the human acceptance statistics for four representative source groups. Acceptance rates range from 93.33\% to 100.00\%, indicating that candidate proposal is generally accurate, but still benefits from manual checking to remove residual mismatches.

\begin{table}[t]
\centering
\small
\setlength{\tabcolsep}{6pt}
\resizebox{\linewidth}{!}{%
\begin{tabular}{lrrrr}
\toprule
Source & \#Candidates & Accepted & Rejected & Acceptance \\
\midrule
Gu Donggao & 899 & 899 & 0 & 100.00\% \\
Kong Yingda & 5,286 & 5,179 & 107 & 97.98\% \\
Du Yu & 5,373 & 5,266 & 107 & 98.01\% \\
L\"u Zuqian & 360 & 336 & 24 & 93.33\% \\
\bottomrule
\end{tabular}
}
\caption{Acceptance rates for later-commentary alignments (LLM candidate proposal + human verification).}
\label{tab:audit_commentary_alignment}
\end{table}

\paragraph{Typical rejection patterns.}
Rejected candidate alignments mainly arise from three sources. First, some later commentaries refer to the correct historical period but to an overly broad textual span, making the proposed match imprecise. Second, some candidates are semantically similar to the target event but mismatch key participants, event roles, or action focus. Third, some compressed headings or summaries are ambiguous enough that multiple canonical passages appear plausible, in which case we conservatively reject the alignment unless a human annotator can verify a unique and appropriate match.

\paragraph{Takeaway.}
Across both curation stages, the role of the LLM is restricted to efficient candidate proposal. Dataset quality is controlled by manual verification and supported by explicit audit statistics. We therefore view the resulting benchmark as a historically grounded, human-verified dataset rather than an AI-generated or synthetic resource.

        

	\end{CJK}
\end{document}